\pdfoutput=1
\documentclass[11pt]{article}

\usepackage[preprint]{acl}

\usepackage{times}
\usepackage{latexsym}
\usepackage{amsmath}
\usepackage{float}
\usepackage[T1]{fontenc}
\usepackage[utf8]{inputenc}

\usepackage{microtype}

\usepackage{inconsolata}

\usepackage{graphicx}
\usepackage{siunitx}
\usepackage{booktabs}
\usepackage{fvextra}

\usepackage{multirow} % provisional add by ZE

\usepackage{tcolorbox}
\tcbuselibrary{breakable}
\usepackage{verbatim}
\usepackage{fvextra} % extends fancyvrb with wrapping
\usepackage{amssymb}
\fvset{
  breaklines=true,        % enable line wrapping
  breakanywhere=true,     % allow breaks inside long tokens
  breaksymbolleft=\tiny\textcolor{gray}{\(\hookrightarrow\)}\,, % wrap marker
  breaksymbolindent=0pt,
  obeytabs=true,
  tabsize=2,
  showspaces=false,
}

\title{WER is Unaware: Assessing How ASR Errors Distort Clinical Understanding in Patient Facing Dialogue}

\author{
  \textbf{Zachary Ellis\textsuperscript{1}\thanks{Equal contribution.}},
  \textbf{Jared Joselowitz\textsuperscript{1}\footnotemark[1]},
  \textbf{Yash Deo\textsuperscript{2}},
  \textbf{Yajie He\textsuperscript{1}},
\\
  \textbf{Anna Kalygina\textsuperscript{1}},
  \textbf{Aisling Higham\textsuperscript{1,3}},
  \textbf{Mana Rahimzadeh\textsuperscript{4}},
  \textbf{Yan Jia\textsuperscript{2}},
\\
  \textbf{Ibrahim Habli\textsuperscript{2}},
  \textbf{Ernest Lim\textsuperscript{1,2}}
\\
\\
  \textsuperscript{1}Ufonia Limited,
  \textsuperscript{2}University of York,
  \textsuperscript{3}Oxford University Hospitals,
  \textsuperscript{4}Moorfields Eye Hospital
\\
  \small{
    \textbf{Correspondence:} \href{mailto:jj@ufonia.com}{jj@ufonia.com}
  }
}

\begin{document}
\maketitle
\begin{abstract}
As Automatic Speech Recognition (ASR) is increasingly deployed in clinical dialogue, standard evaluations still rely heavily on Word Error Rate (WER). This paper challenges that standard, investigating whether WER or other common metrics correlate with the clinical impact of transcription errors. We establish a gold-standard benchmark by having expert clinicians compare ground-truth utterances to their ASR-generated counterparts, labeling the clinical impact of any discrepancies found in two distinct doctor-patient dialogue datasets. Our analysis reveals that WER and a comprehensive suite of existing metrics correlate poorly with the clinician-assigned risk labels (No, Minimal, or Significant Impact). To bridge this evaluation gap, we introduce an LLM-as-a-Judge, programmatically optimized using GEPA to replicate expert clinical assessment. The optimized judge (Gemini-2.5-Pro) achieves human-comparable performance, obtaining 90\% accuracy and a strong Cohen’s $\kappa$ of 0.816. This work provides a validated, automated framework for moving ASR evaluation beyond simple textual fidelity to a necessary, scalable assessment of safety in clinical dialogue.

\end{abstract}

\section{Introduction}
\begin{figure*}[t]
  \centering
\includegraphics[width=\textwidth,height=\textheight,keepaspectratio]{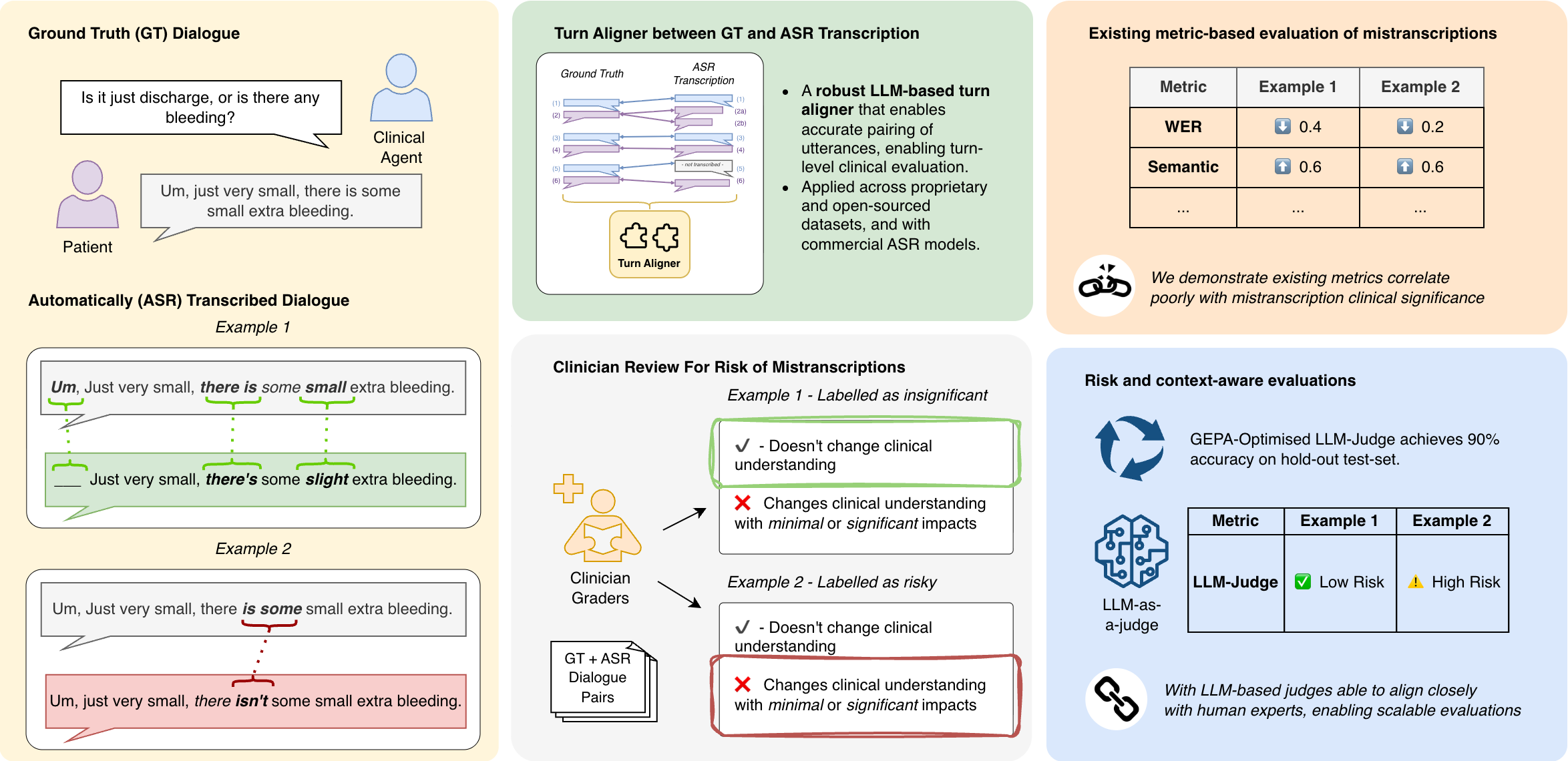}
  \caption{\textbf{Overview of the clinical impact evaluation framework.} \textbf{Left:} Two examples of ASR errors in patient utterances. \textbf{Middle:} We curate a dataset of clinical dialogues and transcriptions, and apply a novel semantically-aware sentence alignment pipeline to enable contextual clinical evaluation. Expert clinicians annotate a dataset of these errors based on our defined scale, labelling the minor change (Ex. 1) as "Insignificant" but the clinically dangerous negation (Ex. 2) as "Impactful". \textbf{Right:} Existing metrics like WER and other semantic scores correlate poorly with clinical risk. Our GEPA-optimized LLM-as-a-Judge closely matches clinical expert ratings.}
  \label{fig:fig1}
\end{figure*}

Patient-facing clinical dialogue agents are increasingly being deployed into live clinical environments, automating tasks from documentation to direct consultations \citep{teo2025generative}. Their performance depends critically on Automatic Speech Recognition (ASR), the "ears" of these clinical agents. While significant research has examined text-level hallucinations in generative models \citep{kim2025medical}, the fidelity of the ASR models that feed these models has received far less scrutiny.

ASR systems are typically benchmarked using Word Error Rate (WER). However, WER is context-agnostic and ill-suited for safety-critical dialogue. It treats all word errors equally, failing to distinguish between trivial disfluencies and clinically hazardous substitutions. For example, a substitution that changes "there is some extra bleeding" to "there isn't some extra bleeding" minimally affects WER yet inverts clinical meaning. Even modern semantic metrics such as BLEURT or BERTScore remain blind to such risks, rewarding textual similarity while ignoring potential clinical consequences.

This paper argues that ASR evaluation in clinical dialogue must evolve towards assessing real clinical impact. To bridge this gap, we make three core contributions (Also illustrated in Figure \ref{fig:fig1}):

% First, we defined a three-point scale for the clinical impact of an ASR mistranscription. We had expert clinicians annotate mistranscriptions from two distinct doctor-patient dialogue datasets; one from a proprietary agent-to-patient dataset of real-world consultations and another from an open-source primary care corpus. This led to a diverse gold-standard annotated dataset of ASR errors from two commercial ASR systems. 

% Second, to enable a precise, turn-by-turn comparison, we found that traditional alignment methods fail to reliably map ground-truth and ASR utterances due to segmentation differences and semantic ambiguity. We therefore developed a robust LLM-based semantic aligner (Sec 5.1), a fundamental contribution essential for assessing downstream clinical harm beyond aggregate WER.

% Finally, we leverage this dataset and tooling to demonstrate that WER and a full suite of modern semantic metrics correlate poorly with expert-assigned clinical impact. We then introduce and validate a scalable solution, by programmatically optimizing an LLM-as-a-judge using GEPA\citep{agrawal2025gepareflectivepromptevolution}. Our resulting judge (Gemini-2.5-Pro) replicates expert clinical assessment with human-comparable performance, obtaining 90\% accuracy and a Cohen's $\kappa$ of 0.816, providing a novel, validated and scalable framework for evaluating the clinical safety of ASR errors. 

\textbf{A clinician-annotated benchmark for ASR clinical impact.} We define a three-point scale for clinical distortion and recruit expert clinicians to annotate mistranscriptions from two doctor–patient datasets; one proprietary and one open-source spanning two ASR systems (Google Chirp and Deepgram Nova-3), yielding a diverse, high-quality dataset of clinically rated ASR errors.
    
\textbf{A robust LLM-based turn aligner.} We outline that traditional alignment methods fail under inconsistent segmentation and semantic ambiguity across ASR providers. Our LLM aligner reasons jointly over meaning, context, and sequence, ensuring accurate pairing of ground-truth and ASR utterances for turn-level comparison.
    
\textbf{A validated LLM-as-a-judge for context-sensitive clinical risk assessment.} Using our dataset, we show that WER and existing semantic metrics correlate poorly with expert-assigned clinical impact. We then optimize an LLM-based evaluator (Gemini-2.5-Pro) via GEPA, achieving 90\% accuracy (Cohen’s $\kappa$ of 0.816), human-comparable performance for scalable clinical safety evaluation.

These contributions provide a concrete step towards risk-informed, context-sensitive evaluations for the development of safer clinical dialogue systems. 

\section{Background and Related Works}
\subsection{Limitations of WER in Clinical Contexts}
The standard metric in ASR evaluation, WER, is fundamentally limited for safety-critical domains like clinical dialogue \citep{SemaScore2024}. As a context-agnostic measure of lexical fidelity (substitutions, deletions, insertions) \citep{RethinkingASREval}, WER overlooks semantic accuracy which is critical in clinical settings where a single misrecognized negation or medication name can reverse meaning and cause severe clinical harm despite a low WER \citep{SemaScore2024}.
Moreover, ASR models optimized for specific benchmarks often show substantially higher WERs in conversational or multi-speaker contexts \citep{RethinkingASREval}, revealing domain variability and the continued need for post-editing for clinical transcriptions. These findings underscore the need for evaluation methods that capture not just textual accuracy but also preservation of clinical meaning.

\subsection{Beyond Lexical Fidelity: Semantic and Hybrid ASR Evaluation}

To overcome the limitations of lexical fidelity, recent work focuses on \textit{semantic fidelity} - measuring the meaning-level distance between reference and hypothesis texts. Early embedding-based methods like \textit{Semantic Distance} \citep{kim2021semantic} use vector representations to quantify similarity, demonstrating better alignment with human perception of quality than WER. 

% More sophisticated hybrid metrics integrate both error quantification and semantic scoring:

% \begin{itemize}
%     \item \textbf{Clinical BERTScore} \citep{ClinicalBERTScore2023} performs utterance-level analysis validated against clinician preferences, showing improved performance over WER in specialized clinical, non-conversational settings.
%     \item \textbf{SeMaScore} \citep{SemaScore2024} combines word level error rate with segment-wise semantic similarity, correlating more closely with expert human assessments, even under noisy or atypical speech conditions \citep{SemaScore2024}.
% \end{itemize}

More sophisticated hybrid metrics integrate both error quantification and semantic scoring. \textbf{Clinical BERTScore} \citep{ClinicalBERTScore2023} conducts utterance-level analysis validated against clinician preferences, showing improved performance over standard WER in specialized, non-conversational settings. Similarly, \textbf{SeMaScore} \citep{SemaScore2024} combines error rates with segment-wise semantic similarity, yielding stronger correlations with expert judgments, even in noisy speech.

These semantic and hybrid metrics show stronger correlation with human judgments of \textit{intelligibility} and \textit{correctability}. Metrics like Human Perceived Accuracy and integrated weighted combinations (e.g., phonetic, semantic, NLI features) achieve better correlations than WER for these domains \citep{HPA2013, IntegratedMetric2025}. 

While these metrics indicate progress, they prioritize semantic resemblance rather than clinical impact. Changes that alter symptom severity ("some mild pain" → "no mild pain") have vastly different clinical implications yet metrics like cosine similarity may fail to capture the consequences of these differences. Thus, even meaning-aware metrics may fail to distinguish clinically consequential errors from inconsequential ones. Furthermore, pre-trained embeddings are robust to ASR errors when predicting user ratings of dialogue systems, suggesting limited sensitivity to clinically relevant ASR distortion \citep{georgila2024}. As a result, existing metrics (whether edit-distance, n-gram, or semantic) remain poor proxies for clinical impact, as they measure textual divergence rather than its effect on clinical understanding or decision-making.

To address this gap, our work moves beyond linguistic similarity toward clinically aware ASR evaluation: quantifying the magnitude of clinical distortion caused by transcription errors. We empirically test how traditional metrics from these three families (edit-distance, overlap-based, and semantic) align with expert clinical judgments and propose an LLM-based evaluator that better reflects the actual clinical consequences of misrecognition.

\subsection{Limitations of Traditional Methods for Sentence-Level Alignment}
Accurately pairing ground-truth clinical utterances with their ASR outputs is essential for valid evaluation, yet conventional alignment methods often fail under the messy, overlapping conditions of real-world dialogue and the inconsistent segmentation produced by different ASR systems.

\textbf{Timestamp-Proximity Alignment.} Simple proximity matching pairs each gold utterance with the nearest ASR hypothesis, but noisy or drifting timestamps often cause swapped or merged segments \citep{bain2023whisperx}. Dynamic Time Warping mitigates rate differences by minimizing cumulative time distance, yet non-monotonic or inaccurate timestamps yield locally coherent but semantically incorrect alignments, especially when utterances are split or merged \citep{Jiang2020}.

\textbf{Text-Based Alignment.} Edit-distance algorithms like Needleman–Wunsch \citep{Needleman1970} and Smith–Waterman \citep{Smith1981} align tokens by lexical similarity alone, ignoring timing and context but they fail when sentence boundaries diverge or ASR outputs contain paraphrases and disfluencies, producing unstable or crossing alignments \citep{snover2006ter}.

\textbf{Embedding-Based Similarity.} 
Sentence embeddings (e.g., \textit{SBERT}) align semantically similar utterances \citep{reimers-gurevych-2019-sentence}, but short backchannels (e.g., "yes", "okay") merge into similar vectors, domain-specific terms are under represented \citep{zheng2021doespretraininghelpassessing}, and ignoring sequence order allows semantically plausible yet positionally inconsistent matches \citep{Liu2022}.

These limitations motivate a context-aware approach that integrates semantic and sequential reasoning. We therefore introduce an LLM-based aligner that robustly maps utterances across fragmented or merged ASR outputs, forming a reliable foundation for our clinical impact benchmark.

\subsection{Large Language Models as Judges for Clinical Impact}

The \textbf{LLM-as-a-Judge} framework provides a promising solution for providing nuanced and contextually aware evaluations of transcription quality, moving beyond the limits of static metrics \citep{gu2024survey, pulikodan2025approach}. LLMs have been leveraged to assess the severity and nature of transcription errors, a process that is essential for operational risk monitoring. Domain-specific adaptations, like \textit{Significant ASR Error Detection (SASRED)} \citep{harvill2024significant}, classify errors as \textit{Significant} (altering key entities or actions) or \textit{Non-Significant} (minor surface changes on an Amazon Alexa general diaogue dataset). However, whilst showing promise with non-expert human evaluators ~\cite{li2024llms}, these model have limitations and variable validation for expert tasks, particularly in healthcare \citep{szymanski2025limitations}.

% LLM-based evaluation frameworks tailored for healthcare assess critical dimensions aligned with clinical workflow standards, including:
% \begin{itemize}
%     \item \textbf{Correctness}: Factual accuracy compared to ground truth.
%     \item \textbf{Helpfulness/Utility}: Practical value, clarity, and actionability of the transcription in a clinical setting.
%     \item \textbf{Logical Coherence}: Ensuring clinical validity and absence of inconsistencies.
% \end{itemize}
% These dimensions allow the LLM judge to function as an automated clinical impact risk assessment tool, flagging high-risk transcripts for mandated human review \citep{LLMJudgeHealthcareAWS}.

These LLM-based frameworks are often tailored to assess critical healthcare dimensions such as factual correctness, clinical utility, and logical coherence to ensure outputs are safe and align with clinical workflow standards \cite{croxford2025automating}. Building on these advances, our methodology tasks expert clinicians, and a subsequent LLM judge, with evaluating transcription errors based on their direct impact on the clinical understanding of a patient's condition, and subsequent risk changes.

\section{Methods}
\subsection{Programmatic Alignment of Ground-Truth and Hypothesis Utterances}
\label{sec:methods_alignment}
To handle segmentation and semantic drift across ASR providers, we employ an LLM-based aligner that performs semantic and structural sentence-level alignment between gold and ASR utterances, instead of relying on time or token matching.

\subsubsection{Prompt Design}

Each conversation contained two ordered sequences: (1) a \textbf{gold transcript} of verified patient utterances with timestamps, and (2) an \textbf{ASR hypothesis} of recognized segments with confidence scores. The LLM aligned each gold utterance $G_i$ to one or more ASR hypotheses $A_j$ under the following constraints: each ASR segment could be matched once; consecutive segment could merge if forming a single utterance; and consecutive gold utterances could map jointly if merged by the recognizer. The model considered semantic similarity, sequential order, and ASR confidence without introducing new text (prompt provided in Appendix~\ref{appendix:llm_aligner_prompt}).

Gemini-2.5-Pro was used with conservative decoding parameters (temperature = 0.1, top-p = 0.95, top-k = 40) to ensure stable long-context outputs (up to 65k tokens). It produced structured JSON alignments specifying indices, match types (\texttt{exact}, \texttt{fuzzy}, \texttt{missing}), and similarity scores.

\subsubsection{Post-Processing and Refinement}
Raw alignments were parsed into structured objects and refined through deterministic rules to ensure validity and robustness: (1) \textbf{duplicate correction} merged consecutive gold segments sharing identical ASR text; (2) \textbf{miss recovery} re-evaluated unmatched gold utterances against unused ASR hypotheses (lexical similarity $\geq$ 0.65); and (3) \textbf{multi-fragment reconstruction} combined gold utterances spanning consecutive ASR fragments, averaging confidence and timestamps.

This hybrid design combines the LLM's reasoning with deterministic corrections, producing content-aware, sequence-consistent alignments resilient to real-world ASR behavior (fragmented or merged outputs). The final alignments, annotated with similarity scores, match types, and multi-fragment indicators, were saved as structured JSON for downstream evaluation. Worked example of the alignment can be seen in Appendix~\ref{appendix:alignment_examples}. 

% todo: fill in appendix

\subsection{Clinician Labelling of Meaning Change and Clinical Impact}
\label{subsec:clinician_labelling}
To evaluate the \textit{clinical impact} of ASR errors, two clinician annotators independently labelled a stratified sample of patient utterances (the Clinical Subset) from post-operative cataract and general-practice consultations. Each example contained a short dialogue segment where only the patient’s final utterance differed between the \textit{ground-truth} and \textit{ASR transcription}. Annotators compared these paired versions and judged whether the transcription error altered the perceived clinical meaning of the exchange. For each instance, clinicians answered the following question:
\begin{quote}
\textit{``If uncorrected, and if you could only read the transcription alone, would it have changed your understanding of the patient’s clinical condition?’’}
\end{quote}

They assigned one of three ordinal labels reflecting the \textbf{magnitude of clinical distortion:}
\begin{itemize}
    \item \textbf{0} – No change in understanding of the patient’s clinical condition
    \item \textbf{1} – Change in understanding with \textit{minimal} clinical impact
    \item \textbf{2} – Change in understanding with \textit{significant} clinical impact
\end{itemize}

Full task instructions and clinician background are outlined in Appendix \ref{appendix:annotation_instructions}. Brief justifications were also recorded to capture reasoning and highlight borderline cases. These annotations formed the reference set for subsequent metric development and correlation analysis.

\subsubsection{Clinician Inter-Annotator Agreement}
\label{sec:iaa}

\begin{figure}[h]
  \includegraphics[width=\columnwidth]{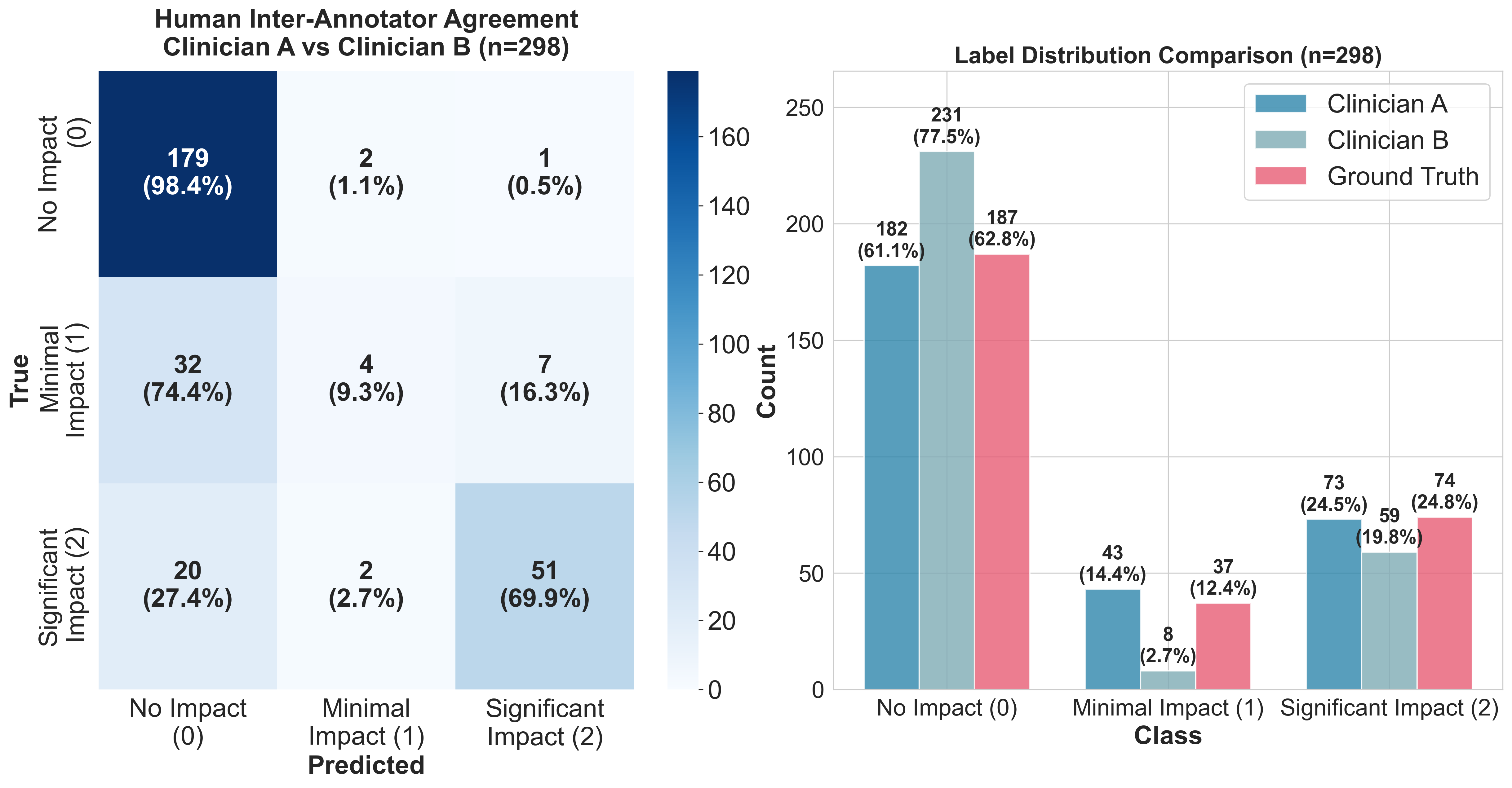}
  \caption{Clinician annotation agreement and final label distribution. \textbf{Left:} IAA between two clinicians on the full labelled subset ($n=298$), with most disagreements between adjacent classes (0 vs.\ 1), yielding 79\% agreement ($\kappa=0.54$). \textbf{Right:} Final adjudicated labels show a predominance of \textit{no-impact} cases, with fewer \textit{minimal} and \textit{significant-impact} examples.}
  \label{fig:human_vs_label_distribution}
\end{figure}

Inter-annotator agreement (IAA) was assessed on the full labelled Clinical Subset using Cohen’s $\kappa$ and raw percentage agreement. Figure~\ref{fig:human_vs_label_distribution} shows the agreement per class and the final adjudicated label distribution. Overall agreement was \textbf{79\%} ($\kappa = 0.54$), indicating moderate agreement. Notably, the majority of disagreements occurred between the 'No Impact' (0) and 'Minimal Impact' (1) classes, highlighting the inherent subjectivity and nuance in distinguishing cosmetic errors from those with minor clinical significance.
Following the initial round, the annotators met to resolve disagreements, producing a reconciled gold-standard set.

%dominated by \textit{no-impact} cases and smaller proportions of \textit{minimal} and \textit{significant-impact} examples for subsequent metric validation. <-- EL Edit, i think figure caption is clear enough. I don't think the "dominated" adds to the narrative. 

\subsection{LLM-as-a-Judge Training}\label{subsec:llm_judge_training}

\textbf{Implementation.}  The LLM judge was implemented using DSPy \citep{khattab2024dspy}, a framework for programmatic prompt optimization. The judge is given a ground truth conversation and and ASR hypothesis as input, and outputs a clinical impact assessment with reasoning.

\textbf{Prompt Optimization via GEPA.} Rather than manually engineering prompts, we used GEPA (Genetic-Pareto) to automatically optimize the judge's instructions \citep{agrawal2025gepareflectivepromptevolution}. GEPA employs a reflective prompt evolution strategy that leverages LLM introspection to iteratively improve prompts based on observed failures.

The optimization process operates as follows: (1) the current prompt is evaluated on minibatches of training examples (batch size = 3); (2) incorrect predictions trigger generation of rich textual feedback describing the narure and severity; (3) a reflection LM uses the current prompt, failed examples, and feedback to generate multiple candidate improved prompts; (4) candidates are evaluated on a validation set and selected via Pareto frontier optimization to maintain diverse high-performing strategies; (5) the process iterates until convergence. Chain-of-Thought approach was used to encourage step-by-step reasoning before classification.

\textbf{Dataset Split and Evaluation Metric.} From the 298 labelled conversation pairs of the Clinical Subset, we created a stratified split of 218 training, 30 validation, and 50 test examples, preserving class distributions.
To encourage clinically meaningful optimization, we used a custom cost matrix $C$ where $C[i,j]$ represents the reward or penalty for predicting class $j$ when the true class is $i$. The matrix heavily penalized missed critical errors (e.g., $C[2,0] = -1.2$) and applied smaller penalties for adjacent-class confusions. Combined with GEPA's textual feedback, this cost-sensitive setup enabled learning of the relative importance of error types. The cost matrix is shown in Appendix~\ref{appendix:gepa_cost_matrix}.
 
% Combined with GEPA's textual feedback, this cost-sensitive setup enabled learning of the relative importance of error types. The full matrix is shown in Appendix~\ref{appendix:gepa_cost_matrix}.

\textbf{Model Configuration.} We used Gemini-2.5-Pro for both executing clinical assessments and generating improved prompts during GEPA's reflection phase. This configuration allowed the model to both perform the judgment task and introspect on its failures to propose improvements. The GEPA optimizer was configured with \texttt{auto='medium'}, Pareto-based candidate selection, and was set to skip examples achieving perfect scores to focus computational resources on challenging cases. The final prompt can be seen in Appendix~\ref{appendix:gepa_final_prompt}.

\subsection{Existing Metric Evaluation}\label{subsec:existing_metrics_eval}
We benchmarked three ASR metric families against clinician-assigned risk labels, using aligned ground-truth and hypothesis pairs. A full list of evaluated metrics is provided in Appendix~\ref{appendix:metric zoo}.

% \begin{itemize}
%     \item \textbf{Edit-Distance Metrics.} Quantify the minimal edits between reference and hypothesis (e.g. WER, Character Error Rate (CER)).
%     \item \textbf{N-gram Overlap Metrics.} These measure lexical overlap using contiguous n-grams (e.g. BLEU, ROUGE).
%     \item \textbf{Learned Semantic Metrics.} These use pretrained neural models to assess meaning preservation (e.g. BERTScore, BLEURT)
% \end{itemize}

\textbf{Edit-distance metrics} (e.g., WER, Character Error Rate (CER)) measure minimal token or character-level edits between reference and hypothesis. The \textbf{N-gram overlap metrics} (e.g., BLEU, ROUGE) capture lexical overlap through contiguous n-grams. Finally, the \textbf{learned semantic metrics} (e.g., BERTScore, BLEURT) use pretrained neural models to assess meaning preservation.

\textbf{Enrichment--delta analysis} was performed, with all scores normalized as \(1 - \text{error rate}\) so higher values indicate better performance. For each metric, we computed the mean score difference between clinically safe (\(y{=}0\)) and high-impact (\(y{=}2\)) transcripts, 
\(\Delta = \mathbb{E}[s \mid y{=}2] - \mathbb{E}[s \mid y{=}0]\), to quantify sensitivity to clinical severity.  This measure captures whether higher-risk cases receive systematically lower quality scores, indicating alignment between metric sensitivity and clinical relevance.

% todo: add: "on the n=278 subset"

%----  Yash i have condensed for now here to get a sense of space ----  % 

%\textbf{Edit-Distance Metrics.} Quantify the minimal edits between reference and hypothesis. We use \textbf{WER}, \textbf{Character Error Rate (CER)}, \textbf{Match Error Rate (MER)}, \textbf{Word Information Lost (WIL)}, and \textbf{S-WER} (a clinically weighted variant).

%\textbf{N-gram Overlap Metrics.} These measure lexical overlap using contiguous n-grams. We compute \textbf{BLEU-1–4} for n-gram precision and \textbf{ROUGE-1}, \textbf{ROUGE-2}, \textbf{ROUGE-L}, \textbf{ROUGE-W}, and \textbf{ROUGE-I} for recall-oriented variants. \textbf{METEOR} extends unigram matching via stemming and synonyms, while \textbf{chrF} and \textbf{chrF++} operate at the character and mixed word–character levels.

%\textbf{Learned Semantic Metrics.} These use pretrained neural models to assess meaning preservation. \textbf{BERTScore} compares contextual embeddings; \textbf{BLEURT}, \textbf{clinical-BLEURT}, and \textbf{BARTScore} predict human-rated or likelihood-based quality; and embedding models \textbf{SBERT-Similarity} and \textbf{SimCSE} use cosine similarity. We also evaluate NLI-based models (\textbf{nli-xsmall}, \textbf{nli-base}, \textbf{nli-large}) framing evaluation as entailment, and task-specific variants (\textbf{sema\_score}, \textbf{intelligibility\_score}, and \textbf{heval\_score}) trained for ASR intelligibility and health-domain assessment.

% ---- % 

% todo: \textbf{Mean Score Differences} - move from results here. Also mention we do Kendall correlation. 

\section{Data}
We curated two complementary datasets of real doctor–patient conversations, differing in domain and ASR provider, to robustly evaluate how ASR mistranscriptions distort clinical meaning.

\subsection{Sources}
Both datasets contain English-language doctor-patient conversations. For the analysis, only the patient’s speech is used.

\textbf{Dora} comprises 21 anonymized production calls from a proprietary telehealth service, \textit{Ufonia Limited} \citep{ufonia2025}, capturing naturally occurring mistranscriptions in a live clinical environment rather than data engineered for this study. These routine post-operative cataract consultations were conducted by an LLM-based conversational agent.

\textbf{Primock57} is an open-source set of 21 mock primary-care dialogues \citep{korfiatis2022primock57}.
\footnote{All accompanying code and the clinician-labelled Primock57 Clinical subsets is publicly released at \url{https://github.com/Ufonia/wer-is-unaware}. The Dora data originates from a proprietary internal dataset and cannot be shared.}

% \footnote{The clinican-labelled Primock57 Clinical and Metrics subsets will be publicly released. The Dora data originates from a proprietary internal dataset and cannot be shared.} 

The combination provides both proprietary and public data analysis, spanning distinct clinical pathways, recording conditions, and ASR providers.

% \subsection{Sources}
% Dora 
% - proprietary calls, anonymized, production, - 21 Dora calls - some known qualitatively to have had clinical impact due to transcription errors, others chosen at random - 21 total.

% Primock 
% - cited, open-sourced, -Primock:
% - 21 conversations chosen at random

% Both with patient-only audio-tracks. 

% both with patient only audio tracks, clinical DoPaCos

% \subsection{Transformation}
% - Dora: 
% --GT transcription carried out on a small subset of the overall dataset 
% -- google legacy ASR used to transcribe calls as they would be in realtime

% --Available GT transcripts were transformed from native format to match DOra format. 
% -- Deepgram transcription used

% Aligner used on both - decomposed the datasets into a series of GT patient utterances paired with preceding doctor utterance, and asr hypotheses matched to utterances.

% Stats of both datasets of 21 calls to report - IN APPENDIX:
% - Total number of patient GT and hyp words across the 21 calls
% - total number of patient gt utterances across the 21 calls
% - Avg number of patient gt and hyp words per call
% - avg number of patient utterances per call
% - avg number of patient words per utterance
% - for Dora, sex and age group distributions of patients over the 21 calls (query the Call table for PatientID, then the Patient table to get age groups (available for all), and sex (available for most - can maybe listen to the rest / access the call audio))

\subsection{Transformation}
All audio was transcribed to produce ground-truth (GT) references and corresponding ASR hypotheses for utterance-level comparison.

\textbf{Ground-Truth Transcription.}
For \textit{Dora}, GT transcripts were created using an human–AI pipeline shown to yield fast, high-accuracy transcriptions \citep{liu-etal-2022-wanli,yuan2021synthbio}. Gemini-2.5-Pro generated initial transcripts from patient audio, which human annotators then verified and corrected. For \textit{Primock57}, we used the provided human transcriptions as GT. In both datasets, adjacent utterances by the same speaker were concatenated into a single, continuous turn.

\textbf{Automatic Transcription.}
To capture variation across commercial systems, \textit{Dora} audio was transcribed using Google Chirp and \textit{Primock57} using Deepgram Nova-3, reflecting diversity in ASR output and segmentation behaviors. Due to cost and time constraints, only one ASR provider was used per dataset. Systematic cross-provider comparisons on the same data are left to future work.
% todo: find better place for this:
% \footnote{Due to cost and time constraints, only one ASR provider was used per dataset. Systematic cross-provider comparisons on the same data are left to future work}

% To reflect variability in commercial ASR systems, \textit{Dora} audio was transcribed using Google Chirp and \textit{Primock57} using Deepgram Nova-3, capturing differences in transcription and segmentation behavior. Owing to cost and time constraints, a single ASR provider was used per dataset; systematic cross-provider comparisons on the same data are left to future work.

\textbf{Utterance Alignment.}
Each dataset was decomposed into aligned pairs of patient GT utterances and ASR hypotheses using the LLM-based semantic aligner (Section~\ref{sec:methods_alignment}), ensuring consistent pairing despite provider-level segmentation differences. For each target utterance, the preceding two doctor turns and the most recent patient turn were appended to preserve conversational context for later clinical annotation. Summary statistics, including WER distribution and average utterance length, are provided in Appendix~\ref{appendix:data_dist}.

\subsection{Curation}\label{subsec:data_curation}
Aligned patient utterances were curated into a clinician-labelling sample. After text cleaning (Appendix~\ref{appendix:dataset_cleaning}), WER was computed between ground-truth and ASR pairs, and perfect matches ($WER = 0$) were excluded. Random sampling from both datasets ensured diversity across speakers and call types. To achieve a balanced range of transcription quality, utterances with higher error rates ($WER \in [0.4, 1)$) were selectively included from \textit{Primock57}. Each pair was manually checked for correct alignment, and any misaligned examples were removed. This curation process yielded a \textbf{Clinical Subset} dataset of 298 examples which was used for clinician labelling (Sec.~\ref{subsec:clinician_labelling}) and the training and testing of the LLM Judge (Sec.~\ref{subsec:llm_judge_training}). 

For the existing metrics evaluation (Sec.~\ref{subsec:existing_metrics_eval}), we additionally filtered out non-lexical tokens, as detailed in Appendix~\ref{appendix:dataset_cleaning}. Twenty of the 298 Clinical Subset pairs became perfect matches ($WER=0$) differing only by these tokens, and were excluded from the existing metrics evaluation (Sec.~\ref{sec:existing_metrics_eval}); this yielded the \textbf{Metrics Subset}. The statistics of both subsets are provided in Table~\ref{tab:data_summary}.

% For the existing metrics evaluation (Sec.~\ref{subsec:existing_metrics_eval}), we further removed non-lexical tokens as detailed in Appendix~\ref{appendix:dataset_cleaning}. Twenty of the 298 pairs became perfect matches ($WER=0$) differing only by these tokens and were excluded from this evaluation. 

% The remaining 278 utterances (121 from Dora and 157 from Primock57) comprised the \textbf{Metrics Subset} for this analysis, with statistics shown in Table~\ref{tab:data_summary}.

% old table 1
% \begin{table}[h]
%     \centering
%     \small
%     \resizebox{\columnwidth}{!}{%
%     \begin{tabular}{lrrrr}
%     \toprule
%     \textbf{Dataset} & \textbf{\# Calls} & \textbf{\# Utterances} & \textbf{Avg. Words/Utt.} & \textbf{Avg. WER} \\
%     \midrule
%     Dora        & 21 & 121 & 9.03 & 0.53 \\
%     Primock57   & 21 & 157 & 12.52 & 0.51 \\
%     \midrule
%     \textbf{Total} & \textbf{42} & \textbf{278} & \textbf{10.99} & \textbf{0.52} \\
%     \bottomrule
%     \end{tabular}
%     }
%     \caption{Datasets used in this study. Statistics are computed using the cleaned patient transcripts on the Metrics Subset (Sec.~\ref{subsec:data_curation}).}
%     \label{tab:data_summary}
% \end{table}

% updated table 1
\begin{table}[h]
    \centering
    \small
    \resizebox{\columnwidth}{!}{%
    \begin{tabular}{llrrrr}
    \toprule
    \textbf{Subset} & \textbf{Source} & \textbf{\# Calls} & \textbf{\# Utterances} & \textbf{Avg. Words/Utt.} & \textbf{Avg. WER} \\
    \midrule
    \multirow{3}{*}{Clinical} 
    & Dora & 21 & 123 & 9.28 & 0.51 \\
    & Primock57 & 21 & 175 & 12.7 & 0.50 \\
    & \textbf{Total} & \textbf{42} & \textbf{298} & \textbf{11.29} & \textbf{0.51} \\
    \midrule
    \multirow{3}{*}{Metrics} 
    & Dora & 21 & 121 & 9.03 & 0.53 \\
    & Primock57 & 21 & 157 & 12.52 & 0.51 \\
    & \textbf{Total} & \textbf{42} & \textbf{278} & \textbf{10.99} & \textbf{0.52} \\
    \bottomrule
    \end{tabular}
    }
    \caption{Datasets used in this study. Statistics are shown for both the Clinical and Metrics subsets; non-lexical tokens are filtered only for the Metrics Subset.}
    \label{tab:data_summary}
\end{table}

% todo: Statistics for the n=278 subset used for the existing metrics evaluation. Statistics are computed using the cleaned patient transcripts.

\section{Results}
\subsection{LLM-Based Aligner}
To ensure the validity of our downstream clinical impact analysis, we first evaluated the LLM-based alignment system. The accuracy of this component is critical, as alignment errors would invalidate the comparisons made by clinician annotators.

\subsubsection{Gold-Standard Alignment Dataset} A human annotator manually aligned patient utterances from a subset of 13 conversations; 7 transcribed with Google Chirp and 6 with Deepgram Nova-3. The dataset contains 463 ground-truth utterances and 445 ASR hypotheses, with each gold utterance mapped to its correct ASR counterpart(s). The annotator labeled one-to-one, one-to-many (merges), many-to-one (splits), and zero-to-one (missed) mappings. This dataset served as the gold standard for the alignment evaluations.

\subsubsection{Evaluation Metrics} We evaluate the LLM-based transcript aligner using two complementary metrics.
\textbf{Classification Accuracy} assesses whether the aligner correctly identifies if an utterance has a corresponding segment in the other transcript. For each of the 463 ground-truth and 445 ASR utterances, this is treated as a binary classification task: correctly labeling an utterance as matched or unmatched (missed). Errors include false positives (incorrectly labeling a match as a miss) and false negatives (failing to detect a true miss).
\textbf{Structural Alignment Accuracy} provides a stricter, mapping-level evaluation. It measures the percentage of ground-truth utterances that were mapped to the \textit{exact} same ASR utterance index (or indices) as specified in the gold-standard annotation. This metric is sensitive to structural errors like boundary shifts, mis-merges, or the incorrect use of a duplicate ASR segment (see Appendix~\ref{appendix:alignment_metrics_example} for a worked example).

\subsubsection{Performance Results}
\paragraph{The LLM-based aligner achieved high, system-agnostic performance.}

\begin{figure}[h]
  \includegraphics[width=\columnwidth]{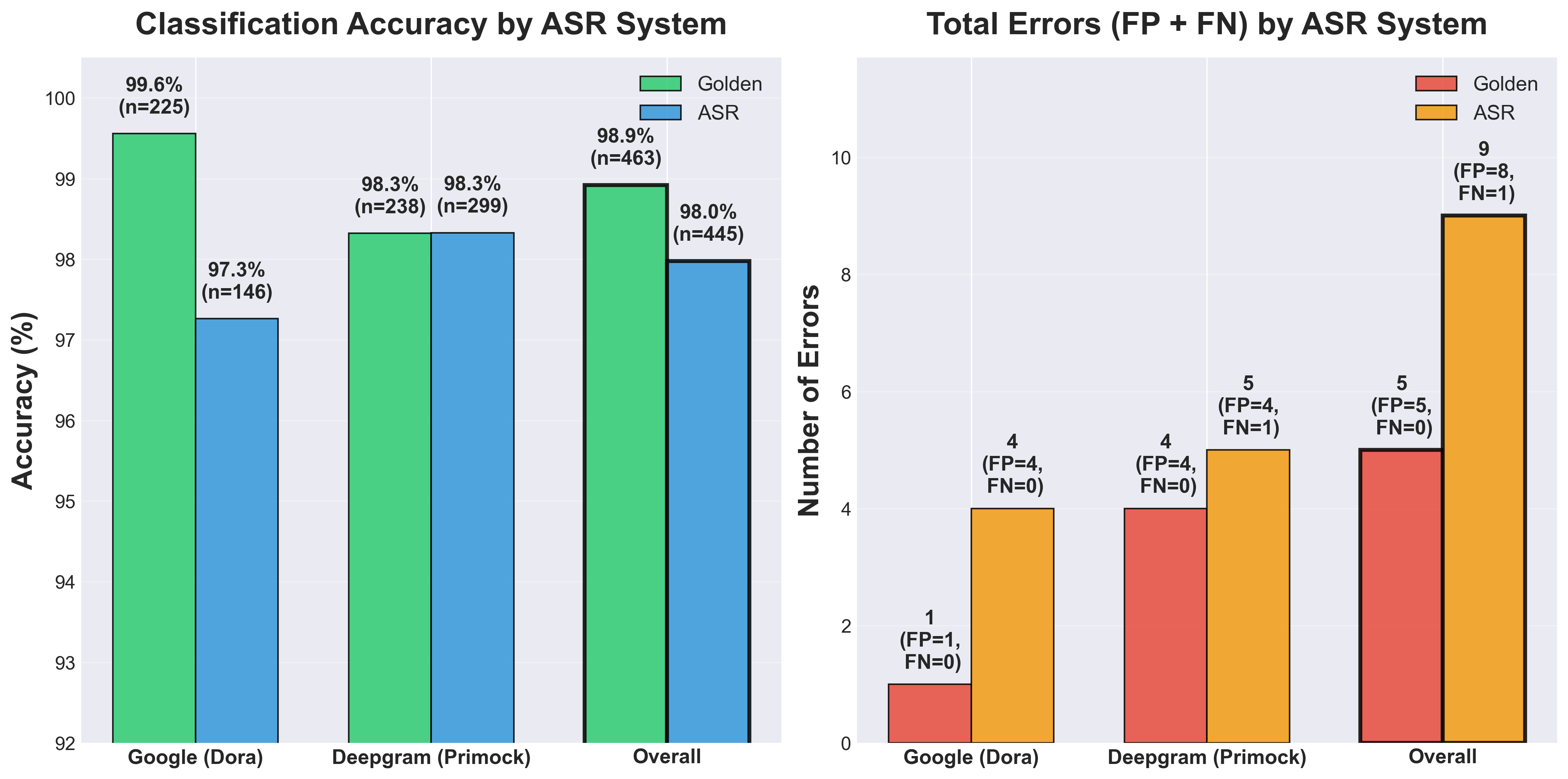}
  \caption{Performance of the LLM-based transcript aligner across Google (Dora) and Deepgram (Primock) ASR hypotheses. The figure shows high classification accuracy ($>98\%$) and low total error counts for both golden and ASR utterances.}
  \label{fig:aligner-asr_comparison}
\end{figure}

For \textit{Classification Accuracy}, results were 98.9\% on gold utterances and 98.0\% on ASR utterances (Figure~\ref{fig:aligner-asr_comparison}). Misclassifications were minimal with one false unmatched case on the golden side for Google, and five for Deepgram, with similarly low counts for ASR results.

% The corresponding total error counts show only minimal misclassifications. The aligner produced one false unmatched case on the golden side for Google and five for Deepgram, with similarly low counts on the ASR side (four errors for each system). These small differences largely reflect the greater utterance fragmentation observed in Deepgram outputs, which occasionally required multi-fragment merging.

For \textit{Structural Alignment Accuracy}, performance remained strong (96.4\%). Minor discrepancies stemmed from boundary drift in long utterances or duplicated ASR fragments, none of which affected clinical meaning in downstream comparisons.

% \paragraph{Structural Alignment Accuracy.} Extended the analysis by verifying whether each golden utterance was mapped to the exact ASR segment(s) in the human-annotated reference, capturing structural inconsistencies such as duplicate ASR usage, boundary shifts, or mismerged utterances.

% Across all cases, the average structural alignment accuracy was 96.4\%, with 6 of 13 cases achieving perfect correspondence. The remaining minor discrepancies were due primarily to boundary mismatches or duplicated ASR fragments within longer utterances.

% Overall, these results demonstrate that the proposed LLM-based aligner generalises well across ASR systems with differing segmentation behavior, achieving near-human accuracy and maintaining structural coherence even without reliable timestamp information.

Overall, these results confirm that the LLM-based aligner is robust and accurate enough, providing a reliable foundation for subsequent clinical impact annotation.

\subsection{Existing Metric Evaluation}
\label{sec:existing_metrics_eval}

% A collection of automated evaluation metrics were calculated on the clinical subset; full details on the implementation and conventional use of these metrics is laid out in Appendix ???. Non-lexical tokens\footnote{The Speechmatics collection of ??? non-lexical tokens such as "uhm", "ahh", and "hmm", was used for this task: INSERT HYPERLINK} were filtered out of the ground truth and transcript for each patient utterance, which were further processed as described in appendix ???, prior to calculation. This filtering of non-lexical tokens led to 20 utterances from the clinical subset having WER = 0; these 20 utterances were excluded from the correlation analysis described here, which was performed on 121 utterances from Dora and 157 utterances from Primock57.

% Kendall's, $\tau$, and Spearman's, $\rho$, rank correlation coefficient were calculated between the reconciled gold-standard set of ordinal clinician labels for each utterance, and each of the calculated automated evaluation metrics. The results in Table~\ref{} show ...

\begin{figure}[h]
  \includegraphics[width=\columnwidth]{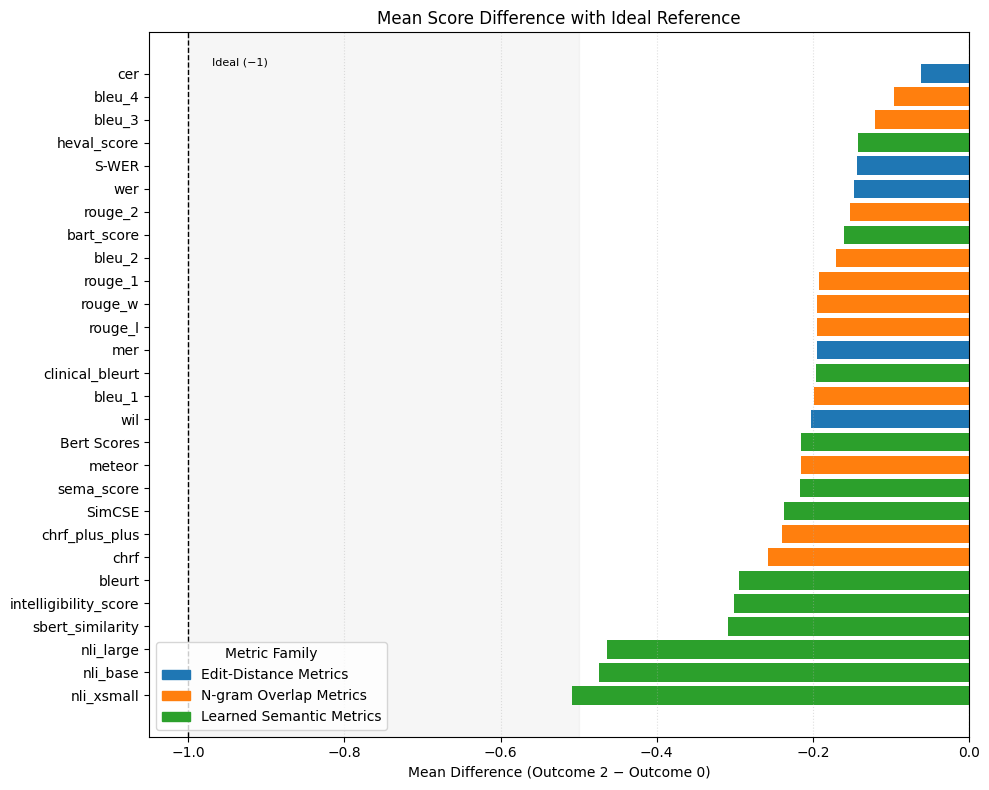}
  \caption{Mean score difference per metric on the \textit{Metrics Subset}, coloured by family; more negative bars indicate stronger alignment with clinical severity.}
  \label{fig:enrichment_delta}
\end{figure}

\paragraph{Existing metrics correlate poorly with clinical labels for risk.} Across all metrics, the enrichment--delta analysis (Figure~\ref{fig:enrichment_delta}) revealed that score differences between high--impact (\(y{=}2\)) and safe (\(y{=}0\)) transcriptions were generally small, confirming that most conventional text metrics only weakly track clinically meaningful errors. Among families, \textit{learned semantic metrics} (e.g., BLEURT, SBERT, NLI models) showed the strongest and most consistent alignment with clinical risk, with clearer score separation between safe and high-impact cases and more negative enrichment deltas, indicating that lower scores generally corresponded to higher clinical severity. \textit{Edit-distance metrics} (WER, CER, etc.) exhibited moderate but less stable associations, while \textit{N-gram overlap metrics} (BLEU, ROUGE, METEOR) provided the weakest discrimination, with high overlap in scores across all clinical categories. A table of results, a Kendall correlation, and an example qualitative error analysis are provided in Appendix \ref{appendix:metrics_detailed}.

Overall, the results suggest that while semantic metrics are relatively better proxies for clinical reliability, no existing metric family reliably reflects real clinical impact, underscoring the need for domain-aware evaluation frameworks.

% todo: add: "on the n=278 subset"

\subsection{LLM-as-a-Judge Automation}

\paragraph{Gemini-2.5-Pro outperforms other state-of-the-art LLMs.} To validate our model selection, we benchmarked the final GEPA-optimized prompt across a suite of leading open source and proprietry models. The full comparison, detailed in Appendix~\ref{appendix:llm_judge_comparison}, shows that Gemini-2.5-Pro achieved a mean Macro F1 of \textbf{0.825} and Cohen's $\kappa$ of \textbf{0.790}. This establishes a clear performance advantage over all other tested models, justifying its use.

\begin{figure}[h]
  \includegraphics[width=\columnwidth]{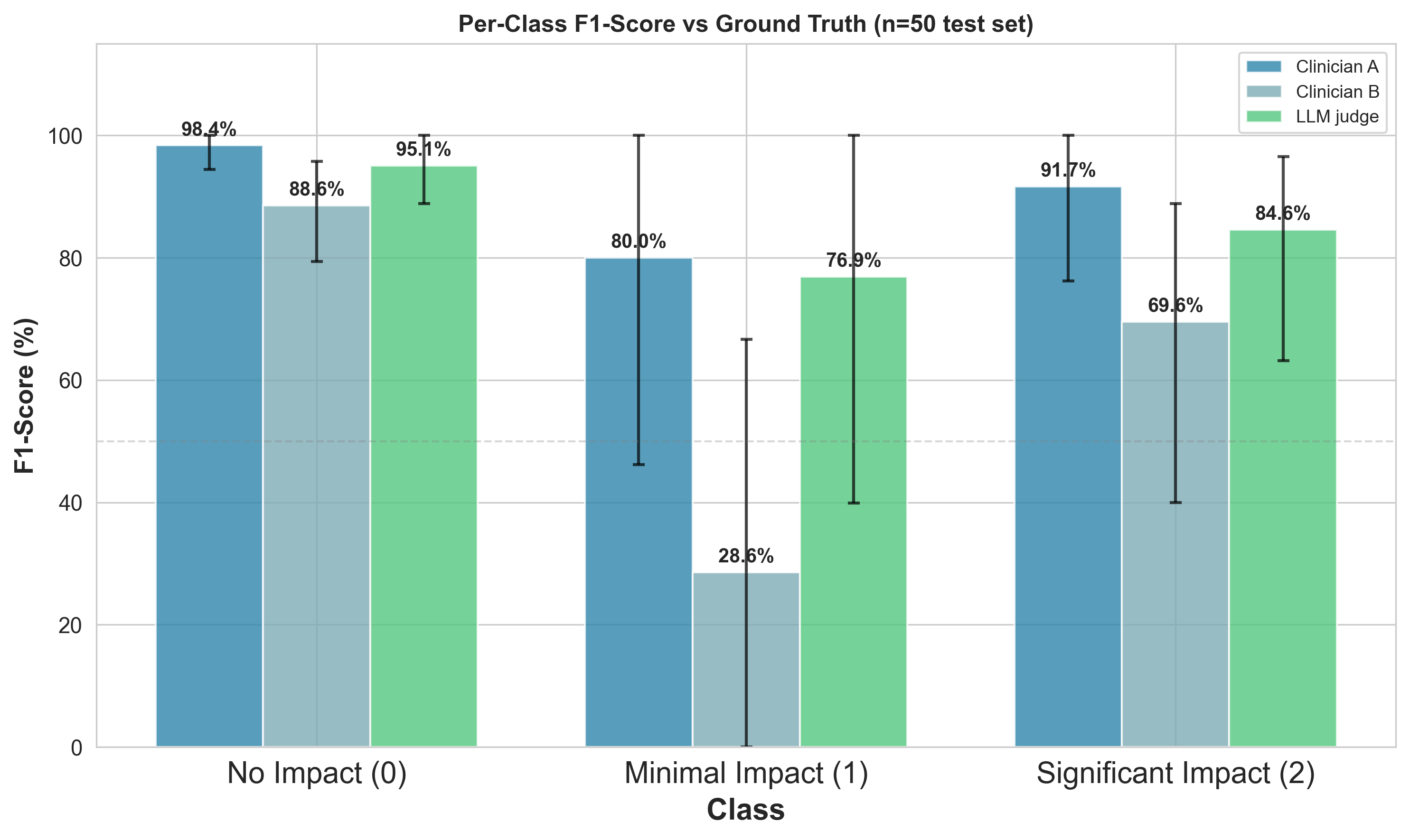}
  \caption{Per class Test set results of clinicians and judge. 95\% confidence interval estimated via 1,000 bootstrap iterations}
  \label{fig:per_class_f1_comparison}
\end{figure}

\paragraph{The Judge excels in ambiguous cases.} A granular per-class F1 analysis (Appendix~\ref{appendix:llm_judge_comparison}) reveals that while most models perform adequately on clear-cut 'No Impact' (Class 0) or 'Significant Impact' (Class 2) cases, they consistently fail on the nuanced 'Minimal Impact' (Class 1) category. This reflects the difficulty of this class, which also proved most challenging for human annotators (Figure~\ref{fig:per_class_f1_comparison}). Gemini-2.5-Pro was the only model to achieve an F1 score > 0.5 (it got 0.655) for this difficult class, demonstrating a superior capacity for nuanced clinical assessment (Figure~\ref{fig:run_f1_score_per_class}).

% \paragraph{The LLM Judge reaches human-comparable performance on the test set.} From Table~\ref{tab:judge_stats_comparison}, on the 50 held-out cases, the Judge's \textbf{90\%} accuracy with \(\mathbf{\kappa=0.816}\) puts it directly between the two expert annotators (Clinician A: 94\%, $\kappa=0.891$; Clinician B: 80\%, $\kappa=0.567$). This demonstrates that the automated judge is not just accurate as compared to other models, but also comparable to human experts on this task.

\paragraph{Judge achieves human-comparable performance and agreement.} The LLM Judge's performance is comparable to its human expert counterparts. From Table~\ref{tab:judge_stats_comparison}, its \textbf{90\%} accuracy (\(\mathbf{\kappa=0.816}\)), places it between the two expert annotators (Clinician A: 94\%; Clinician B: 80\%).

Furthermore, from Table~\ref{tab:iaa_with_judge}, the Judge's agreement patterns mirrors expert reliability. Its pairwise $\kappa$ with Clinician A (0.713) and Clinician B (0.497) is consistent with the inter-clinician $\kappa$ of 0.505. This demonstrates the Judge successfully operates within the same range of expert subjectivity.

% Table of average and correlation
\begin{table}[h]
  \centering
  \resizebox{\columnwidth}{!}{%
  \begin{tabular}{lccc}
    \hline
     \textbf{Comparison} & \textbf{Acc (95\% CI)} & \textbf{Cohens $\kappa$ (95\% CI)} \\
    \hline
    LLM Judge vs Gold    & 90\% [82.0-96.0]  & 0.816 [0.649-0.933]          \\
    Clinician A vs Gold  & 94\% [88.0-100.0]  & 0.891 [0.764-1.000]         \\
    Clinician B vs Gold  & 80\% [68.0-90.0] & 0.567 [0.336-0.767]      \\\hline
  \end{tabular}
  }
  \caption{Agreement with gold-standard labels across 50 cases with 95\% confidence interval estimated via 1,000 bootstrap iterations. The LLM Judge shows high alignment with human clinicians.}
  \label{tab:judge_stats_comparison}
\end{table}

% % Table of agreement between clinicians and judge
% \begin{table}[h]
%     \centering
%     \resizebox{\columnwidth}{!}{%
%     \begin{tabular}{lccc}
%     \hline
%     & \textbf{Clinician A} & \textbf{Clinician B} & \textbf{Judge} \\
%     \hline
%     \textbf{Clinician A} & --- & 0.505 (0.285, 0.708) & \textbf{0.713 (0.535, 0.867)} \\
%     \textbf{Clinician B} & 0.505 (0.285, 0.708) & --- & 0.497 (0.273, 0.702) \\
%     \textbf{Judge} & \textbf{0.713 (0.535, 0.867)} & 0.497 (0.273, 0.702) & --- \\\hline
%     \end{tabular}
%     }
%     \caption{Agreement between clinicians and judge using Cohen's $\kappa$ with 95\% bootstrap confidence intervals.}
%     \label{tab:iaa_with_judge}
% \end{table}

\paragraph{The Judge mirrors the stronger clinician across classes, with greatest uncertainty on the minority class.}
Beyond aggregate scores, the Judge’s per-class F1 performance (Figure~\ref{fig:per_class_f1_comparison}) closely tracks that of the stronger human annotator. The Judge achieves \textbf{95.1\%} on \emph{No Impact}, \textbf{76.9\%} on \emph{Minimal Impact}, and \textbf{84.6\%} on \emph{Significant Impact}, compared with Clinician~A’s 98.4\% / 80.0\% / 91.7\% and Clinician~B’s 88.6\% / 28.6\% / 69.6\%. Both the Judge and Clinician~A perform nearly perfectly on clear-cut \emph{No Impact} cases, show moderate decline on \emph{Significant Impact}, and exhibit the greatest variability on the ambiguous \emph{Minimal Impact} class, reflecting its inherent difficulty.

% Figure of per class F1

% \paragraph{Judge–clinician agreement is comparable to clinician–clinician agreement.}
% From Table~\ref{tab:iaa_with_judge}, pairwise Cohen’s \(\kappa\) shows Judge–Clinician~A agreement of \(\kappa=0.713\) \([0.535,0.867]\), comparable to Clinician~A–Clinician~B at \(\kappa=0.505\) \([0.285,0.708]\); Judge–Clinician~B agreement is \(\kappa=0.497\) \([0.273,0.702]\), reflecting the same range of variability observed among human experts. Therefore, the Judge’s agreement patterns are consistent with inter-clinician reliability.

\section{Discussion}
Our findings highlight the critical gap between existing ASR evaluation and clinical safety. We demonstrate that metrics must move beyond textual fidelity (e.g. WER), and even semantic fidelity, both insufficient proxies for risk for clinical dialogue tasks, and therefore falling short for required safety evidence for regulated medical devices \citep{teferra2017iso}. Our LLM-judge closes this gap by achieving human-comparable performance in a challenging and nuanced task, using anonymized real-world data from production use of a conversational system, supported by an open-sourced primary care dataset.

Furthermore, programmatic optimization via GEPA yields not only a high-performing judge but also a reproducible, auditable artefact for the prompt tuning process. Unlike manual prompt engineering, GEPA's training process enables alignment with best-practice AI governance requirements in medicine, established for more traditional ML systems \citep{gallifant2025tripod, ganapathi2022tackling}. Additionally, this analysis was enabled by the LLM-based sentence aligner, which ensured robust utterance-level pairing between ground-truth and ASR transcripts despite segmentation drift or merged utterances.

Limitations include the benchmark's moderate size ($n=298$), and the initial focus on a smaller set of clinical domains. Future work should expand the evaluation to more clinical pathways and involve a larger, more diverse group of clinical labellers.

% Limitations include the benchmark’s moderate size ($n=298$) and its initial focus on a limited set of clinical domains. Future work should expand the evaluation to additional clinical pathways and involve a larger, more diverse group of clinical annotators.

%Our results demonstrate that standard ASR metrics are unreliable proxies for clinical safety. Both WER and modern semantic metrics show weak correlation with clinician-assigned risk, confirming that lexical correctness does not guarantee clinical safety. This gap underscores the need for evaluation methods that reflect clinical meaning rather than surface accuracy. Our optimized LLM Judge, trained via GEPA, bridges this gap by achieving human-comparable performance (90\% accuracy, $\kappa=0.816$) and outperforming both state-of-the-art LLMs and a human annotator on the \textit{Minimal Impact} class. This validated, automated framework provides a scalable tool for ASR use in clinical applications.

%Limitations include the benchmark’s moderate size ($n=298$), the inherent subjectivity of clinical judgment reflected in inter-annotator agreement, and the focus on low-acuity domains (cataract and primary care). Future work should extend to higher-risk specialties.

\section{Conclusion}
Standard ASR evaluation fails to ensure patient safety. We show empirically that existing metrics like WER are insufficient, and introduce an expert-annotated benchmark and a validated LLM Judge that achieves human-level accuracy in classifying clinical risk. Together, these contributions establish the first scalable framework for certifying the clinical safety of transcription systems in conversational clinical dialogues, enabling their responsible development and deployment in healthcare.

% Standard ASR evaluation fails to ensure patient safety. We show empirically that existing metrics like WER are inadequate, and introduce an expert-annotated benchmark plus a validated LLM Judge achieving human-level accuracy in clinical risk classification. Together, these form the first scalable framework for certifying transcription safety in clinical dialogues, enabling responsible healthcare deployment.

% As ASR becomes integral to clinical workflows, evaluation must evolve from measuring lexical fidelity to assessing clinical safety. We provide empirical evidence that existing metrics, including WER, are insufficient for this task, and introduce an expert-annotated benchmark and a validated LLM Judge that achieves human-level accuracy in classifying clinical risk. Together, these contributions establish the first scalable framework for certifying the clinical safety of ASR systems, enabling their responsible development and deployment in healthcare.

\bibliography{custom}

\appendix

\onecolumn

\section{LLM Aligner Prompt}
\label{appendix:llm_aligner_prompt}

\begin{tcolorbox}[title={Prompt for aligning sentences from groundtruth to ASR hypothesis},
  breakable, colback=gray!5, colframe=gray!60!black, boxrule=0.3pt, arc=2pt]
\begin{Verbatim}
You are an expert at aligning speech transcripts. I need you to match patient utterances from a golden transcript with ASR (Automatic Speech Recognition) hypothesis results.

{{golden_text}}
{{asr_text}}

TASK: Align each golden transcript utterance (G0, G1, etc.) with the most appropriate ASR result(s) (A0, A1, etc.). 

RULES:
1. You can only use utterances that exist in the input - DO NOT create new text
2. Each golden utterance should be matched to one ASR result, multiple ASR results, or marked as missing
3. Each ASR result can only be used ONCE - no ASR result should appear in multiple alignments
4. Make reasonable fuzzy matches even if the text isn't perfect - ASR often has errors
5. Consider semantic similarity, temporal proximity, and confidence scores
6. Multiple consecutive ASR results can be combined to match one golden utterance if they represent fragments
7. IMPORTANT: If an ASR result contains content that spans multiple consecutive golden utterances, those golden utterances should ALL be matched to that same ASR result

EXAMPLE of rule 7:
- Golden G5: "I know I understand that"
- Golden G6: "but it's different with the cataract"
- ASR A3: "I know I understand that but it's different with the cataract"
- CORRECT: G5→A3, G6→A3 (both use same ASR)
- WRONG: G5→missing, G6→A3 (creates artificial missing)

OUTPUT FORMAT (JSON):
{
  "alignments": [
    {
      "golden_index": 0,
      "asr_indices": [0],
      "match_type": "exact|fuzzy|missing",
      "similarity_score": 0.95,
      "explanation": "Brief reason for this alignment"
    },
    ...
  ]
}

Provide only the JSON response, no other text.
\end{Verbatim}
\end{tcolorbox}

\twocolumn

\section{Worked Alignment Examples and Alignment Quality Metrics}
\label{appendix:alignment_examples}
\subsection{Worked Alignment Examples Across Common ASR Segmentation Scenarios}

ASR systems produce variable-length output segments that do not correspond reliably to linguistic sentences or speaker turns. As a result, alignment is performed between contiguous groups of golden utterances and contiguous groups of ASR segments, allowing one-to-one, many-to-one, one-to-many, and partial matches.

Below we illustrate the most common alignment scenarios observed in practice. \\

\textbf{Scenario A: One-to-one alignment (clean segmentation).}

\noindent
Golden utterance (index 0): \\
\textit{``Hello, good morning.''}

\noindent
ASR segment (index 0): \\
\textit{``hello good morning''}

\noindent
Alignment:
\begin{itemize}
  \item \texttt{golden\_indices = [0]}
  \item \texttt{asr\_indices = [0]}
  \item \texttt{match\_type = exact}
\end{itemize}

This represents the ideal case where ASR segmentation aligns cleanly with the golden transcript. \\

\textbf{Scenario B: Many-to-one alignment (ASR splits a single utterance).}

\noindent
Golden utterance (index 2): \\
\textit{``Yes. Uh, my name is John Smith. And I was born on the fifth of April, uh, nineteen seventy three.''}

\noindent
ASR segments (indices 2--3):
\begin{itemize}
  \item \textit{``yes my name is john smith''}
  \item \textit{``i was born on the fifth of april nineteen''}
\end{itemize}

\noindent
Alignment:
\begin{itemize}
  \item \texttt{golden\_indices = [2]}
  \item \texttt{asr\_indices = [2, 3]}
  \item \texttt{match\_type = multi\_fragment}
\end{itemize}

This occurs when a single semantic utterance is split across multiple ASR chunks. \\

\textbf{Scenario C: One-to-many alignment (golden utterance spans multiple lines).}

Golden utterance (indices 8--9):
\begin{itemize}
  \item \textit{``Um it's much more like itchy. And my eczema was more like only in the arm. But now it's also on the chest. And in the on the, on the hands as well.''}
  \item \textit{``Like pretty, yeah when, like I like, for instance hiking, during the weekend. And I am, I can't really do it anymore, because it's like very like I wanted to do that last weekend. And because super painful, and I I have to take like showers every day to be able to, cannot uh ease this itching part, which is very very annoying.''}
\end{itemize}

\noindent
ASR segments (indices 11): \\
\textit{``it's much more like itchy and my eczema was more like only in the arms and now also on the chest and in the in the on the hands as well like yeah when like i like for instance hiking during the weekends and i i can't really do it anymore because it's like very like like i wanted to do that last weekend and it was super painful and i i have to take like showers every day to be able to kind of ease this itching part which is very very annoying''}

\noindent
Alignment:
\begin{itemize}
  \item \texttt{golden\_indices = [8, 9]}
  \item \texttt{asr\_indices = [11]}
  \item \texttt{match\_type = multi\_fragment}
\end{itemize}

Here, a golden sentence that was broken over two lines (due to the doctor's interruption), was transcribed as a single sentence by the ASR provider.

\subsection{Worked Example of Alignment Evaluation Metrics}
\label{appendix:alignment_metrics_example}

We illustrate the computation of the two alignment evaluation metrics using a small synthetic example.

\paragraph{Toy example setup.}
Consider a short patient interaction with the following golden utterances:

\begin{itemize}
    \item G0: ``Yes, my name is John Smith.''
    \item G1: ``I was born on the fifth of April.''
    \item G2: ``I have some swelling on my elbow.''
    \item G3: ``Bye.''
\end{itemize}

The corresponding ASR output is segmented as:

\begin{itemize}
    \item A0: ``yes my name is john smith''
    \item A1: ``i was born on the fifth of april''
    \item A2: ``i have some swelling''
    \item A3: ``on my elbow''
    \item A4: ``thank''
\end{itemize}

The gold-standard alignment specifies:
\begin{align*}
    (0) &\rightarrow [0] \\
    (1) &\rightarrow [1] \\
    (2) &\rightarrow [2, 3]
\end{align*}

with unmatched content:
\[
\text{Unmatched golden} = \{3\},
\text{Unmatched ASR} = \{4\}.
\]

An LLM aligner produces the following prediction:
\begin{align*}
    (0) &\rightarrow [0] \\
    (1) &\rightarrow [1] \\
    (2) &\rightarrow [2]
\end{align*}

with predicted unmatched sets:
\[
\text{Unmatched golden} = \{3\},
\text{Unmatched ASR} = \{3, 4\}.
\]

This prediction contains a structural error (missing ASR index $3$) and an unmatched-detection error (incorrectly marking ASR index $3$ as unused).

\paragraph{Classification Accuracy.}
Classification Accuracy evaluates whether each utterance or ASR segment is correctly labeled as matched or unmatched.

\textbf{Golden utterances:}
\[
\text{TP} = 1,\;
\text{FP} = 0,\;
\text{FN} = 0,\;
\text{TN} = 3,
\]
yielding an accuracy of:
\[
\frac{TP + TN}{TP + TN + FP + FN} = \frac{4}{4} = 1.00.
\]

\textbf{ASR segments:}
\[
\text{TP} = 1,\;
\text{FP} = 1,\;
\text{FN} = 0,\;
\text{TN} = 3,
\]
yielding an accuracy of:
\[
\frac{TP + TN}{TP + TN + FP + FN} = \frac{4}{5} = 0.80.
\]

This reflects that the aligner correctly identifies unused golden content, but incorrectly discards one ASR segment that should have been aligned.

\paragraph{Structural Alignment Accuracy.}
Structural Alignment Accuracy evaluates whether the aligner produces the \emph{exact} same ASR index grouping for each golden utterance as in the gold-standard alignment.

In this example, the alignments for G0 and G1 are correct, while the alignment for G2 is incorrect due to a missing ASR fragment. With four golden utterances in total, this yields:
\[
\text{Structural Accuracy} = \frac{4 - 1}{4} = 0.75.
\]

\paragraph{Interpretation.}
Classification Accuracy captures whether the aligner correctly determines which content should participate in alignment, while Structural Alignment Accuracy captures whether the resulting fragment groupings are structurally correct. Both metrics are required to fully characterize alignment quality in the presence of inconsistent ASR segmentation.

\section{Annotation Instructions}
\label{appendix:annotation_instructions}
The following show the full annotation instructions that clinician labellers used for initial labelling, and subsequent disagreement resolution. Clinician annotators' backgrounds are also reported here. 

\subsection{Demographic Backgrounds for Clinician Annotators}

\begin{table}[h]
\centering
\begin{tabular}{p{0.20\linewidth} p{0.30\linewidth} p{0.35\linewidth}}
\textbf{Clinician Labeller} & \textbf{Years of Post-graduate Experience} & \textbf{Primary Specialty} \\
\hline
A & 8 years & Ophthalmology \\
B & 7 years & Ophthalmology \\
\end{tabular}
\end{table}

Both clinicians were ophthalmologists due to the nature of the Dora dataset being primarily in cataract surgery follow up, with the remainder of the Primock dataset being primary care clinical interactions. 

\subsection{Full annotation instructions provided to labellers}
\textbf{Objective} \\
We are examining how transcription errors can affect clinical meaning in a clinical conversational context, and looking to build better metrics to capture this. 

Your task is to then assess if your understanding of the patient’s clinical situation is different because of the transcription errors, and what the potential clinical impact of that is. 
All the excerpts you will see are from either post-cataract surgery follow-up calls, or general practice consultations. You'll be shown short sections of these dialogues. For each one, you will see the Ground Truth (what the patient actually said) and a Transcription (an automatically transcribed version that may contain errors).
Only the last sentence is different in these transcription vs ground-truth examples.

\vspace{1em}
\textbf{The Annotation Task} \\
For each example, you will be presented with: \\
The patient's Ground Truth utterance (what was actually said) \\
The patient's Transcription utterance (what was transcribed, with only the last sentence being different) \\
Some additional context is provided of the conversation preceding these two statements as well. Please read the two versions of the conversation, and then, consider this question: 

\begin{quote}
“If uncorrected, and if you could only read the transcription-alone, would it have changed your understanding of the patient's clinical condition?”
\end{quote}

Please assign one of the following three labels and provide a brief justification for your choice. \\
0 - It does not change my understanding of the patient’s clinical condition. \\
1 - It changes my understanding of the patient’s clinical condition, with minimal clinical impact. \\
2 - It changes my understanding of the patient’s clinical condition, with significant clinical impact. 

Below shows additional detail for each category and some examples. 

\vspace{1em}
\textbf{Detailed Rating Definitions}

\textbf{0 - It does not change my understanding of the patient’s clinical condition.} \\

This will commonly include situations where \\
The meaning of the utterance is preserved despite the errors \\
The meaning is changed, but not of something relating to the patient’s clinical condition or situation - (e.g. administrative, patient feedback, address or logistical). \\
The ground-truth was not clear, and the transcription didn’t really change it or misrepresent it.

\vspace{1em}
\textbf{1 - It changes my understanding of the patient’s clinical condition, with minimal clinical impact.} \\

This is where something is changed, and it is relevant to the patient’s clinical condition, however, it has either minimal or no clinical significance. This will commonly include situations where \\
A patient preference is missing or misunderstood, but this for something administrative or not extremely clinically risky. \\
A part of an utterance is missed or wrong, and this is clinically relevant, (i.e. does relate to the patient's condition, expectations, treatment plan, family history) but it is unlikely to affect the overall outcomes, or it was not a critical piece of information that was missed or wrong. 

\vspace{1em}
\textbf{2 - It changes my understanding of the patient’s clinical condition, with significant clinical impact.} \\

This is where something is changed, and it is relevant to the patient’s clinical condition, and it potentially leads to significant clinical impact. This will commonly include situations where: \\
A patient answers about a symptom but significant parts of it are altered or omitted. \\
A fact is missed or wrong, especially if it’s clinically relevant to the scenario, and if its meaning has been totally changed. \\
Key history parameters are wrong (e.g. past medical history, drug history, family history) \\
The patient could have had relevant questions or other points that weren't captured. 

\vspace{1em}
\textbf{Examples}

This is a mock-example: \\
Note that in all example pairs only the sentences in bold are different between the ground-truth and context+transcripts. 

\textbf{Example A}

\begin{table}[h]
\centering
\begin{tabular}{p{0.45\linewidth} p{0.45\linewidth}}
\textbf{Context + Ground Truth} & \textbf{Context + Transcript} \\
\hline
(21) Doctor: Is your eye red? & (21) Doctor: Is your eye red? \\
(21) Patient: No & (21) Patient: No \\
(22) Doctor: Great, and Is your eye painful? & (22) Doctor: Great, and Is your eye painful? \\
\textbf{(22) Patient: Well it’s not painful, just a bit gritty.} & \textbf{(22) Patient: Well it’s painful} \\
\end{tabular}
\end{table}

This would be labelled 2 - as going off the transcript alone, my understanding of the situation has completely flipped from “just a bit gritty” and “not painful” to simply “it’s painful”. This is clinically significant as is a core clinical question.

\vspace{1em}
\textbf{Example B}

\begin{table}[h]
\centering
\begin{tabular}{p{0.45\linewidth} p{0.45\linewidth}}
\textbf{Context + Ground Truth} & \textbf{Context + Transcript} \\
\hline
(21) Doctor: Is your eye red? & (21) Doctor: Is your eye red? \\
(21) Patient: No & (21) Patient: No \\
(22) Doctor: Great, and Is your eye painful? & (22) Doctor: Great, and Is your eye painful? \\
\textbf{(22) Patient: Well it’s not painful, just a bit gritty that's all.} & \textbf{(22) Patient: not painful, just bit gritty} \\
\end{tabular}
\end{table}

Although the sentence is notably changed, the meaning between both transcript and ground truth is not changed and so this would be labelled 0. 

\vspace{1em}
\textbf{Example C}

\begin{table}[h]
\centering
\begin{tabular}{p{0.45\linewidth} p{0.45\linewidth}}
\textbf{Context + Ground Truth} & \textbf{Context + Transcript} \\
\hline
(21) Doctor: Okay and do you drink? & (21) Doctor: Okay and do you drink? \\
(21) Patient: No & (21) Patient: No \\
(22) Doctor: Do you smoke? & (22) Doctor: Do you smoke? \\
\textbf{(22) Patient: Um, occasionally, you know, just socially, the odd cigarette. But i don’t vape or anything.} & \textbf{(22) Patient: Um, yeah occasionally, social cigarettes and vape anything.} \\
\end{tabular}
\end{table}

This would be labelled 1 - going off the transcript alone, it sounds like the patient is a social smoker and vapes rather than just a social smoker. However, this is unlikely to be of significant clinical impact overall given this is a social history, and we are able to understand in both that they are a smoker. 

\vspace{1em}
\textbf{Example D}

\begin{table}[h]
\centering
\begin{tabular}{p{0.45\linewidth} p{0.45\linewidth}}
\textbf{Context + Ground Truth} & \textbf{Context + Transcript} \\
\hline
(21) Doctor: Okay and do you drink? & (21) Doctor: Okay and do you drink? \\
(21) Patient: No & (21) Patient: No \\
(22) Doctor: Do you smoke? & (22) Doctor: Do you smoke? \\
\textbf{(22) Patient: um} & \textbf{(22) Patient: ah} \\
\end{tabular}
\end{table}

This would be labelled 0 - Both transcripts are unclear, and the ground truth didn’t misrepresent or edit it. 

\vspace{1em}
\textbf{Clarifying Instructions} \\
You are comparing between the Ground-truth and transcription. If the original ground-truth transcription is confusing or unclear, please assess any additional change in meaning from the Ground Truth.

\section{GEPA Cost Matrix}
\label{appendix:gepa_cost_matrix}
We employed a cost-sensitive metric using matrix C, where $C[i,j]$ is the reward/penalty for predicting class j when the true class is i:

\begin{table}[h]
    \centering
    \caption{Confusion matrix $C$}
    \begin{tabular}{c|ccc}
    \toprule
     & \textbf{Pred 0} & \textbf{Pred 1} & \textbf{Pred 2} \\
    \midrule
    \textbf{True 0} & 1.2 & 0.3 & -1.0 \\
    \textbf{True 1} & 0.3 & 1.5 & 0.5 \\
    \textbf{True 2} & -1.2 & 0.4 & 1.5 \\
    \bottomrule
    \end{tabular}
\end{table}

The matrix encodes three priorities: 
\begin{enumerate}
    \item  reward correct classifications (diagonal > 1.0), 
    \item  heavily penalise missing critical errors ($C[2,0] = -1.2$), 
    \item  tolerate adjacent class confusions (small values for $C[i,i \pm 1]$). 
\end{enumerate}

This reflects that missing a significant transcription error is more dangerous than over-classifying a cosmetic error, while acknowledging that boundaries between adjacent severity levels contain inherent ambiguity. The aggregate score guides GEPA optimization toward clinically safe performance.

\onecolumn

\section{GEPA Final Prompt}
\label{appendix:gepa_final_prompt}

\begin{tcolorbox}[title={Prompt to assess clinical errors},
  breakable, colback=gray!5, colframe=gray!60!black, boxrule=0.3pt, arc=2pt]
\begin{Verbatim}
You are an expert medical analyst. Your task is to assess the clinical impact of errors in an AI-generated transcription of a medical conversation.
You will be given two versions of a conversation:
1.  `ground_truth_conversation`: The accurate, human-verified transcript.
2.  `transcription_conversation`: The machine-generated transcript, which may contain errors.
Your goal is to compare the `transcription_conversation` to the `ground_truth_conversation` and determine if any differences (errors) could negatively impact patient care. You must output your assessment in two parts: a `reasoning` section and a `clinical_impact` classification.

### Core Principle for Assessment
The core principle is to determine if a clinician reading the `transcription_conversation` would have a different understanding or make a different medical decision than if they had read the `ground_truth_conversation`. You must be highly sensitive to any changes in clinically critical information, such as symptoms, medications, measurements, or diagnoses.

### Severity Levels for Clinical Impact
You must classify the impact of the transcription errors into one of three distinct severity levels.

**Class 0: No Clinical Impact**
This is the most common class. Errors in this category are "cosmetic" and do **not** change the clinical meaning of the conversation in any way. A clinician would arrive at the exact same understanding and conclusions from both transcripts.
Assign Class 0 for errors such as:
*   **Filler words:** Omission or addition of words like "uh," "um," "like," "you know," "so."
*   **Punctuation and Capitalization:** Differences in commas, periods, or letter casing.
*   **Minor Rephrasing:** The wording is slightly different, but the core clinical fact remains identical and unambiguous. For example, "probably about six months ago" vs. "that's six months okay" both clearly convey the "six months" timeframe.

**Class 1: Potential for Minor Clinical Impact**
Errors in this category introduce ambiguity, omit non-critical context, or alter a piece of clinical information in a way that *could* lead to a misunderstanding, but is unlikely to cause immediate harm. These errors might prompt a clinician to seek clarification but do not fundamentally alter the patient's current clinical picture.
Assign Class 1 for errors such as:
*   **Ambiguity Requiring Clarification:** The omission of a patient's confirmation that their questions were answered. This leaves the clinician unsure if the patient understands their care plan, which requires follow-up but isn't an immediate safety risk.
*   **Omission of Future Intentions:** The omission of a patient's answer about whether they plan to have surgery on their other eye in the future. This creates a gap in planning but does not affect the diagnosis or treatment of the current condition.
*   **Altered Timelines or Frequencies:** Changing a duration or frequency to a different, but plausible, value (e.g., "two days" becomes "two weeks"; "once a day" becomes "twice a day").
*   **Ambiguous Symptoms:** Making a symptom's description less clear (e.g., "sharp pain" becomes "some pain").

**Class 2: Significant Clinical Impact**
Errors in this category are critical and could directly lead to a wrong diagnosis, an incorrect treatment plan, or a serious adverse event. These errors fundamentally change a key clinical fact related to the patient's current condition, history, or diagnostic process. **Reserve Class 2 for errors that could directly affect diagnosis, treatment, or patient safety.**
Assign Class 2 for errors such as:
*   **Omission/Alteration of Diagnostic Reasoning:** This is a high-priority error. For example, omitting a patient's statement where they explicitly connect their symptoms to a known family history of a specific condition (e.g., "I'm worried this is a migraine... I know it's genetic from my mom and sister"). This information is a critical part of the History of Present Illness (HPI) and directly influences the diagnostic workup. Its omission is a significant loss.
*   **Negation Errors:** Changing a positive to a negative, or vice-versa (e.g., "no chest pain" becomes "chest pain"; "patient is not allergic" becomes "patient is allergic").
*   **Critical Value Errors:** Changing a specific, critical number, such as a medication dosage (e.g., "10mg" becomes "100mg") or a vital sign.
*   **Symptom/Condition Errors:** Introducing a new, incorrect symptom or diagnosis, or omitting a critical one mentioned in the ground truth.
*   **Anatomical Errors:** Changing the location of a symptom (e.g., "left arm" becomes "right arm").

### Your Response Format
Your output must include two components:
1.  **`reasoning`**: Provide a step-by-step analysis.
    *   First, identify the specific, clinically relevant differences between the ground truth and the transcription.
    *   Second, analyze whether these differences alter clinically relevant information (symptoms, medications, timelines, diagnostic reasoning, etc.).
    *   Finally, justify your choice of `clinical_impact` class by explaining how the error would (or would not) affect a clinician's understanding or decision-making, referencing the specific criteria for the class you have chosen.
2.  **`clinical_impact`**: Provide the single integer corresponding to your classification (`0`, `1`, or `2`).
\end{Verbatim}
\end{tcolorbox}

\twocolumn

\section{Evaluation Metrics Comparison}
\label{appendix:metric zoo}
% \begin{table}[h!]
%     \centering
%     \caption{Mean Difference in Score (Condition 2 minus Condition 0) for Various Metrics}
%     \label{tab:mean_diff_metrics}
%     \begin{tabular}{l S[table-format=-1.3]}
%         \toprule
%         \textbf{Metric} & {\textbf{Mean Difference}} \\
%         \midrule
%         WER & -0.148 \\
%         SEMA Score & -0.216 \\
%         Intelligibility Score & -0.301 \\
%         HEVAL Score & -0.142 \\
%         Clinical BLEURT & -0.196 \\
%         BLEURT & -0.294 \\
%         BART Score & -0.160 \\
%         MER & -0.195 \\
%         WIL & -0.202 \\
%         CER & -0.062 \\
%         BLEU-1 & -0.198 \\
%         BLEU-2 & -0.170 \\
%         BLEU-3 & -0.121 \\
%         BLEU-4 & -0.097 \\
%         ROUGE-L & -0.195 \\
%         ROUGE-1 & -0.193 \\
%         ROUGE-2 & -0.152 \\
%         ROUGE-W & -0.195 \\
%         ChrF & -0.257 \\
%         ChrF++ & -0.239 \\
%         METEOR & -0.216 \\
%         SBERT Similarity & -0.309 \\
%         NLI XSmall & -0.508 \\
%         NLI Base & -0.475 \\
%         NLI Large & -0.463 \\
%         BERTScore & -0.215 \\
%         SimCSE & -0.237 \\
%         S-WER & -0.144 \\
%         \bottomrule
%     \end{tabular}
% \end{table}

\subsection{Edit-Distance Metrics}
This family of metrics quantifies the dissimilarity between an ASR-generated hypothesis and a ground-truth reference by calculating the minimum number of edits required to make them identical. They are fundamentally error rates, where a lower score indicates a better transcription.

\begin{itemize}
    \item \textbf{Word Error Rate (WER)} is the de facto standard, measuring word-level substitutions, deletions, and insertions.
    \item \textbf{Character Error Rate (CER)} is a variant of WER that operates at the character level, useful for morphologically complex languages.
    \item \textbf{Match Error Rate (MER)} is a bounded version of WER that includes matches in its denominator, making it less sensitive to reference length.
    \item \textbf{Word Information Lost (WIL)} is an information-theoretic extension of WER that weighs errors based on their probabilistic impact.
    \item \textbf{Semantic-WER (S-WER)} is an enhanced WER that adds semantic weights to penalize errors on important words more heavily \cite{roy2021semanticwerunifiedmetricevaluation}.
\end{itemize}

\subsection{N-gram Overlap Metrics}
Borrowed primarily from the field of machine translation, these metrics evaluate quality by measuring the lexical overlap of n-grams (contiguous sequences of items) between the hypothesis and reference texts.

\begin{itemize}
    \item \textbf{BLEU (Bilingual Evaluation Understudy)} is a precision-focused metric that measures n-gram overlap with a penalty for overly short transcriptions.~\cite{papineni2002bleu}
    \item \textbf{ROUGE (Recall-Oriented Understudy for Gisting Evaluation)} is a recall-focused metric for n-gram overlap; variants include ROUGE-N, ROUGE-L, and ROUGE-W.~\cite{lin2004rouge}. The F-Measure is reported for all ROUGE variants in this work.
    \item \textbf{METEOR (Metric for Evaluation of Translation with Explicit ORdering)} is an advanced metric aligning unigrams using stemming and synonym matching for greater flexibility.~\cite{banerjee2005meteor}
    \item \textbf{chrF and chrF++} compute an F-score based on character n-gram overlap, with the `++` version also including word n-grams.~\cite{popovic2015chrf}
\end{itemize}

\subsection{Learned Semantic Metrics}
This modern class of metrics leverages deep learning models to move beyond lexical overlap and capture semantic similarity, determining if the core \textit{meaning} of the text is preserved.

\begin{itemize}
    \item \textbf{BERTScore} utilizes contextual embeddings to compute a nuanced semantic similarity score between tokens.~\cite{zhang2019bertscore}~\cite{ClinicalBERTScore2023}
    \item \textbf{BLEURT} is a regression-based model trained on human quality ratings to predict the quality of a generated text.~\cite{sellam2020bleurt}
    \item \textbf{ClincalBLEURT} is a version of BLEURT fine-tuned on  family medicine and orthopaedic notes.~\cite{abacha-023-investigation}
    \item \textbf{BARTScore} is a generation-based metric using the BART model to assess quality based on conditional probability.~\cite{yuan2021bartscore}
    \item \textbf{SBERT-Similarity and SimCSE} compute sentence embeddings for the hypothesis and reference and measure their cosine similarity.~\cite{reimers2019sentence}~\cite{gao2021simcse}
    \item \textbf{Natural Language Inference (NLI) Scores} repurpose Natural Language Inference models to measure semantic equivalence using bidirectional entailment (mutual entailment), following the approach in \cite{IntegratedMetric2025}. %\textbf{NLI-based Scores} repurpose Natural Language Inference models to check for logical entailment or contradiction.~\cite{laban2022summac}.
    \item \textbf{HEVAL - Hybrid Evaluation Metric for Automatic Speech Recognition Tasks} introduces a hybrid metric combining traditional error-based scoring (e.g. edit distances) on non-keywords with embedding-based semantic distance for ASR outputs. \citep{sasindran2023h}
    \item \textbf{SeMaScore} combines phonetic error rates with segment-wise semantic similarity to yield stronger correlations with expert judgements in noisy speech settings \citep{SemaScore2024}. In our experiments the word embeddings were extracted using deberta-large-mnli.
    \item \textbf{Intelligibility Score} is a hybrid metric that fuses phonetic, semantic, and NLI-based features to align ASR evaluation with human judgements of comprehensibility. \citep{IntegratedMetric2025}.In our experiments the word embeddings were extracted using RoBERTa-large fine-tuned on SNLI. 
    
\end{itemize}

\clearpage

\onecolumn

\subsection{Evaluation Metrics Model and Implementation Details}

Additional information on the implementation of the different evaluation metrics is provided in Table~\ref{tab:metric_models}.
\begin{table*}[h!]
    \centering
    
    % Use a paragraph column 'p' for the model names to allow text wrapping
    \begin{tabular}{l p{7cm} l}
        \toprule
        \textbf{Metric} & \textbf{Model / Implementation} & \textbf{Source} \\
        \midrule
        \multicolumn{3}{l}{\textbf{Edit-Distance Metrics}} \\
        \midrule
        WER & \multirow{4}{*}{jiwer} & \multirow{4}{*}{\href{https://pypi.org/project/jiwer/}{[Link]}} \\
        CER &  &  \\
        MER &  &  \\
        WIL &  &  \\
        \cmidrule(lr){1-3}
        S-WER & sentence-transformer & \href{https://huggingface.co/sentence-transformers}{[Link]} \\
        \midrule
        \multicolumn{3}{l}{\textbf{N-gram Overlap Metrics}} \\
        \midrule
        BLEU variants & \textit{NLTK} & \href{https://www.nltk.org/_modules/nltk/translate/bleu_score.html}{[Link]} \\
        \addlinespace % Adds a little vertical space
        ROUGE variants & \textit{rouge\_score} & \href{https://github.com/google-research/google-research/tree/master/rouge}{[GitHub]} \\
        \addlinespace
        ChrF(\texttt{++}) & \textit{sacrebleu} & \href{https://github.com/mjpost/sacrebleu/blob/master/sacrebleu/metrics/chrf.py}{[GitHub]} \\
        METEOR & \textit{NLTK} & \href{https://www.nltk.org/api/nltk.translate.meteor_score.html}{[Link]} \\
        \midrule
        \multicolumn{3}{l}{\textbf{Learned Semantic Metrics}} \\
        \midrule
        SEMA Score & \textit{microsoft/deberta-large-mnli} & \href{https://huggingface.co/microsoft/deberta-large-mnli}{[Link]} \\
        Intelligibility Score & \textit{ynie/roberta-large-snli\_mnli\_fever\_anli\_R1\_R2\_R3-nli} & \href{https://huggingface.co/ynie/roberta-large-snli_mnli_fever_anli_R1_R2_R3-nli}{[Link]} \\
        HEVAL Score & \textit{roberta-base} & \href{https://huggingface.co/FacebookAI/roberta-base}{[Link]} \\
        Clinical BLEURT & \textit{bleurt-oss-21 (fine-tuned)} & \href{https://github.com/abachaa/EvaluationMetrics-ACL23/tree/main/ClinicalBLEURT}{[GitHub]} \\
        BLEURT & \textit{bleurt-oss-21} & \href{https://github.com/google-research/bleurt}{[GitHub]} \\
        BART Score & \textit{facebook/bart-large-cnn} & \href{https://github.com/neulab/BARTScore}{[HF Link]} \\
        SBERT Similarity & \textit{all-MiniLM-L6-v2} & \href{https://huggingface.co/sentence-transformers/all-MiniLM-L6-v2}{[HF Link]} \\
        \addlinespace
        NLI XSmall & \textit{cross-encoder/nli-deberta-v3-xsmall} & \href{https://huggingface.co/cross-encoder/nli-deberta-v3-xsmall}{[HF Link]} \\
        NLI Base & \textit{cross-encoder/nli-deberta-v3-base} & \href{https://huggingface.co/cross-encoder/nli-deberta-v3-base}{[HF Link]} \\
        NLI Large & \textit{cross-encoder/nli-deberta-v3-large} & \href{https://huggingface.co/cross-encoder/nli-deberta-v3-large}{[HF Link]} \\
        \addlinespace
        BERTScore & \textit{microsoft/deberta-large-mnl} & \href{https://huggingface.co/microsoft/deberta-large-mnli}{[HF Link]} \\
        SimCSE & \textit{princeton-nlp/sup-simcse-bert-base-uncased} & \href{https://github.com/princeton-nlp/SimCSE}{[Link]} \\
        \bottomrule
    \end{tabular}
    \caption{Model Specifications and Sources for Evaluation Metrics}
    \label{tab:metric_models}
\end{table*}

\clearpage

\twocolumn

\section{Dataset Distribution Details}
\label{appendix:data_dist}
\begin{figure}[h]
  \centering
  \includegraphics[width=\columnwidth]{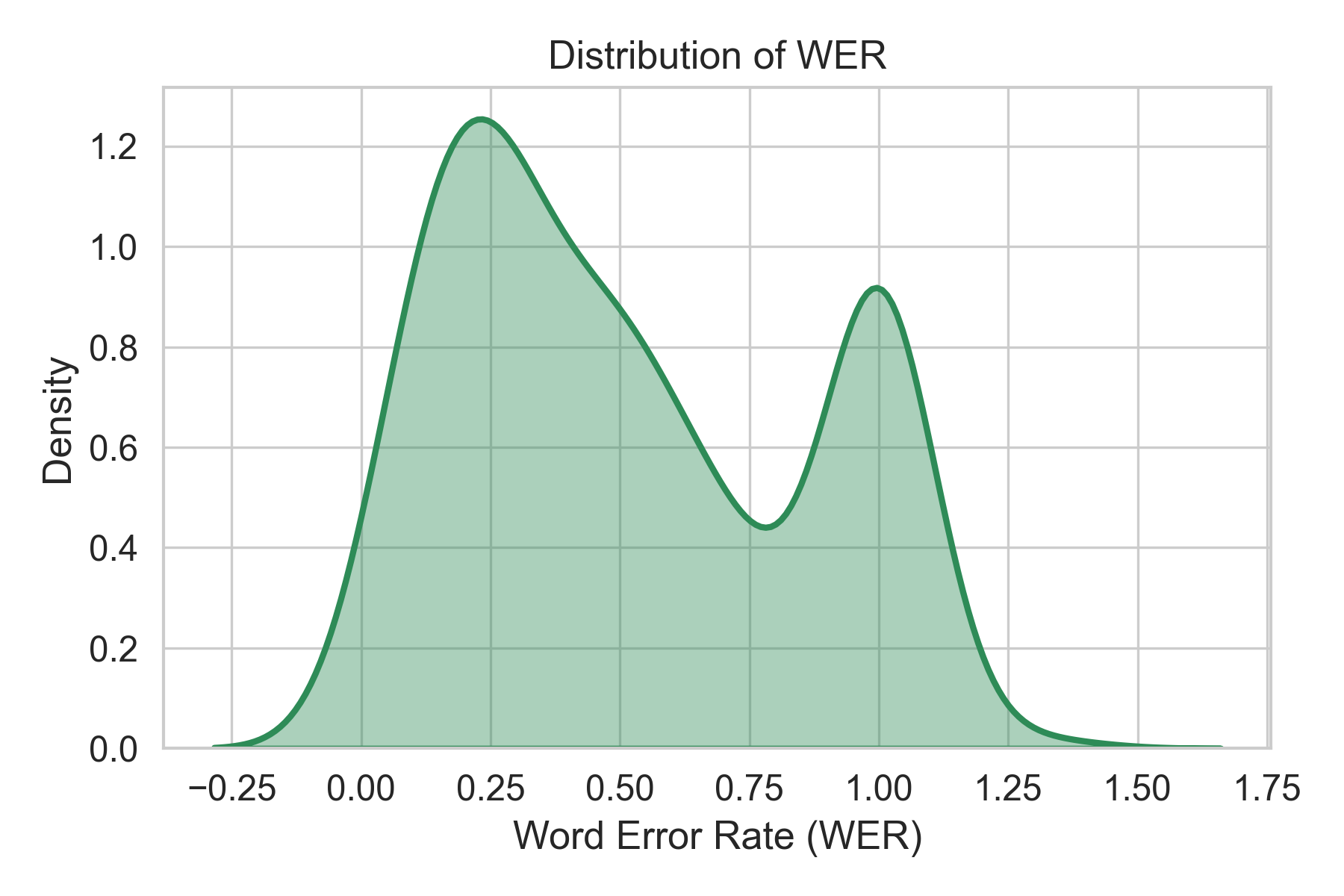}
  \caption{Distribution of WER across utterances on the combined Metrics Subset. A bimodal distribution is observed, with one peak at a low WER and a second smaller peak at high WER.}
  \label{fig:wer_dist}
\end{figure}

\begin{figure}[h]
  \centering
  \includegraphics[width=\columnwidth]{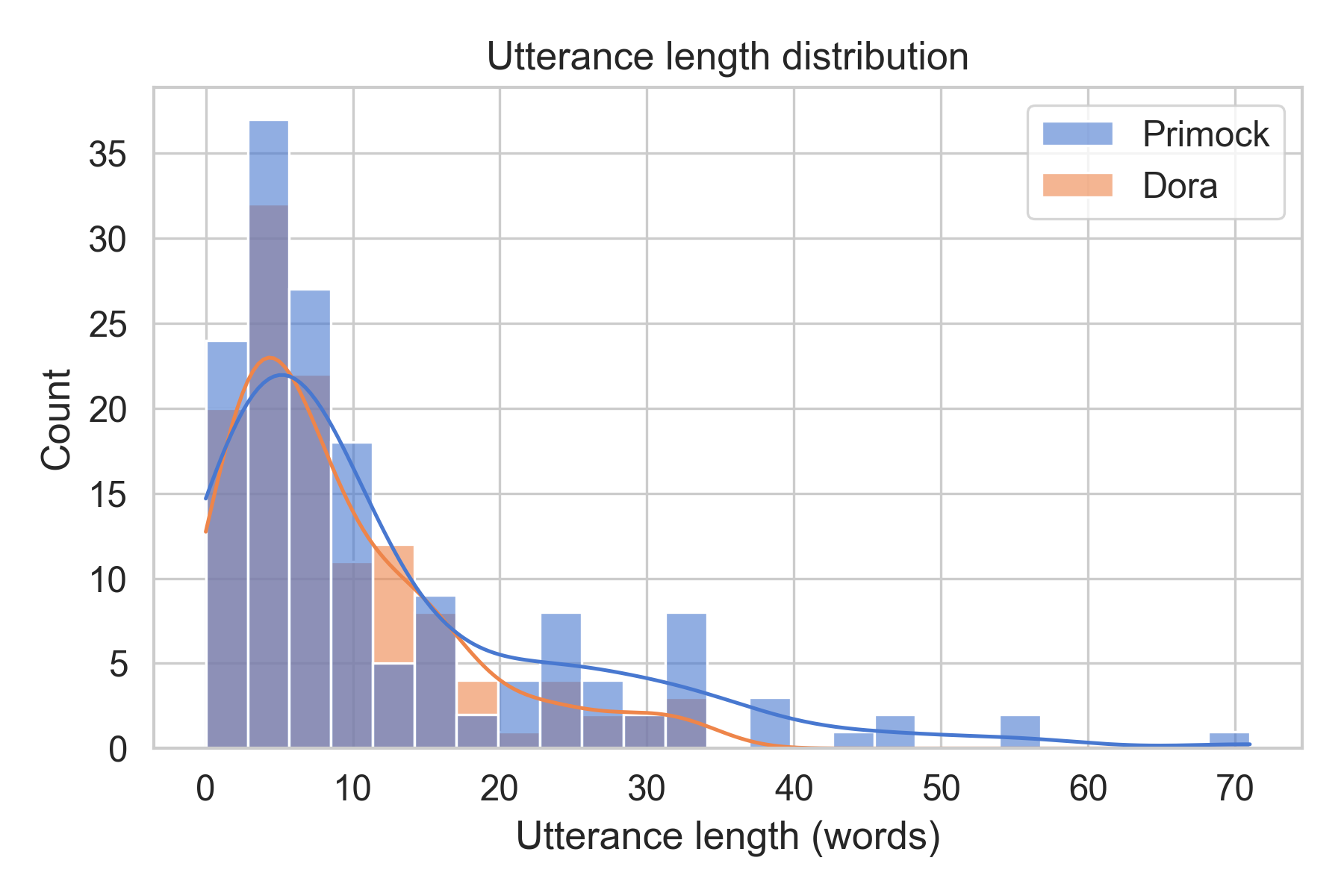}
  \caption{Utterance length distribution for \textit{Dora} and \textit{Primock57} on the Metrics Subset. Both datasets are skewed toward short utterances, with Primock57 showing a longer tail, reflecting occasional extended patient turns.}
  \label{fig:utt_len_dist}
\end{figure}

\section{Dataset Cleaning Process}
\label{appendix:dataset_cleaning}

Prior to metric calculation, both the reference (ground truth) and hypothesis (ASR output) transcripts underwent standardised text normalization to ensure fair comparison. The preprocessing pipeline, implemented using the Python \texttt{jiwer} library, consisted of the following sequential transformations:

\begin{enumerate}
    \item \textbf{Number Normalisation}: All numeric expressions were converted to their word equivalents using British English conventions (e.g., "1st" -> "first", "23" -> "twenty-three") via the \texttt{num2words} library.
    \item \textbf{Case Normalisation}: All text was converted to lowercase.
    \item \textbf{Punctuation Standardisation}: Hyphens were replaced with spaces to prevent word concatenation, and all remaining punctuation was removed.
    \item \textbf{Whitespace Normalisation}: Multiple consecutive spaces were collapsed into single spaces, and leading/trailing whitespace was removed.
    \item \textbf{Non-Lexical Token Removal} (for the Metrics Subset only, Sec.~\ref{subsec:data_curation}): Disfluencies and filler words (e.g., "um", "uh", "hmm") were removed based on a predefined lexicon of 43 non-lexical tokens adapted from Speechmatics documentation \citep{RUSSELL2024100163}.
\end{enumerate}

This preprocessing was applied identically to both reference and hypothesis texts immediately before each metric calculation (WER, BLEU, ROUGE, etc.), ensuring consistent normalisation across all evaluation metrics.

\clearpage
\section{Detailed Results for Existing Metric Evaluation}
\label{appendix:metrics_detailed}
\subsection{Mean Difference Scores per Metric - Enrichment--Delta Analysis}
\begin{table}[h!]
    \centering
    \caption{Mean Difference in Score (Condition 2 minus Condition 0) Grouped by Metric Family}
    \label{tab:mean_diff_metrics_grouped}
    \begin{tabular}{l S[table-format=-1.3]}
        \toprule
        \textbf{Metric} & {\textbf{Mean Difference}} \\
        \midrule
        \multicolumn{2}{l}{\textbf{Edit-Distance Metrics}} \\
        \midrule
        WER & -0.148 \\
        CER & -0.062 \\
        MER & -0.195 \\
        WIL & -0.202 \\
        S-WER & -0.144 \\
        \midrule
        \multicolumn{2}{l}{\textbf{N-gram Overlap Metrics}} \\
        \midrule
        BLEU-1 & -0.198 \\
        BLEU-2 & -0.170 \\
        BLEU-3 & -0.121 \\
        BLEU-4 & -0.097 \\
        \addlinespace % Adds a little vertical space
        ROUGE-L & -0.195 \\
        ROUGE-1 & -0.193 \\
        ROUGE-2 & -0.152 \\
        ROUGE-W & -0.195 \\
        \addlinespace
        ChrF & -0.257 \\
        ChrF++ & -0.239 \\
        METEOR & -0.216 \\
        \midrule
        \multicolumn{2}{l}{\textbf{Learned Semantic Metrics}} \\
        \midrule
        SEMA Score & -0.216 \\
        Intelligibility Score & -0.301 \\
        HEVAL Score & -0.142 \\
        Clinical BLEURT & -0.196 \\
        BLEURT & -0.294 \\
        BART Score & -0.160 \\
        SBERT Similarity & -0.309 \\
        \addlinespace
        NLI XSmall & -0.508 \\
        NLI Base & -0.475 \\
        NLI Large & -0.463 \\
        \addlinespace
        BERTScore & -0.215 \\
        SimCSE & -0.237 \\
        \bottomrule
    \end{tabular}
\end{table}

\subsection{Kendalls Correlation between Metrics and Clinical Labels}

\begin{table}[h!]
    \centering
    \caption{Kendall's $\tau$ Correlation Grouped by Metric Family}
    \label{tab:kendall_tau_grouped}
    \begin{tabular}{l S[table-format=-1.6]}
        \toprule
        \textbf{Metric} & {$\mathbf{\tau}$ \textbf{(Kendall's Tau)}} \\
        \midrule
        \multicolumn{2}{l}{\textbf{Edit-Distance Metrics}} \\
        \midrule
        WER & 0.206765 \\
        CER & 0.232115 \\
        MER & 0.214383 \\
        WIL & 0.215302 \\
        S-WER & 0.227910 \\
        \midrule
        \multicolumn{2}{l}{\textbf{N-gram Overlap Metrics}} \\
        \midrule
        BLEU-1 & -0.218176 \\
        BLEU-2 & -0.188999 \\
        BLEU-3 & -0.149426 \\
        BLEU-4 & -0.125599 \\
        \addlinespace % Adds a little vertical space
        ROUGE-L & -0.224263 \\
        ROUGE-1 & -0.223657 \\
        ROUGE-2 & -0.163319 \\
        ROUGE-W & -0.224263 \\
        \addlinespace
        ChrF & -0.289068 \\
        ChrF++ & -0.261439 \\
        METEOR & -0.235693 \\
        \midrule
        \multicolumn{2}{l}{\textbf{Learned Semantic Metrics}} \\
        \midrule
        SEMA Score & -0.222682 \\
        Intelligibility Score & -0.394572 \\
        HEVAL Score & 0.280065 \\
        Clinical BLEURT & -0.381359 \\
        BLEURT & -0.372970 \\
        BART Score & -0.253080 \\
        SBERT Similarity & -0.323138 \\
        \addlinespace
        NLI XSmall & -0.422054 \\
        NLI Base & -0.389810 \\
        NLI Large & -0.394935 \\
        \addlinespace
        BERTScore & -0.233922 \\
        SimCSE & -0.371572 \\
        \bottomrule
    \end{tabular}
\end{table}

\clearpage

\onecolumn

\subsection{Qualitative Error Analysis of Existing Metrics}
\label{appendix:qualitative_analysis}

Two examples from the Primock57 portion of the Metrics Subset demonstrate specific scenarios where traditional evaluation metrics fail to detect clinical distortion. Table \ref{tab:qual_example_a} (Example A) presents a high-risk scenario where the ASR system mistranscribed a key word "throat" as "so". Expert clinicians assigned this a \textbf{Significant Impact (2)} rating, noting that the error transforms a specific symptom denial, "not throat", into the vague phrase, "not so". In the context of assessing breathing difficulties, this ambiguity is dangerous as it fails to clearly rule out the symptom, potentially affecting patient treatment. However, many standard metrics, whether edit-distance, n-gram overlap-based, or learned semantic, failed to capture this dangerous distortion. Table \ref{tab:metrics_comparison} shows that the transcript had a relatively low WER (0.1176) and high scores across embedding-based similarity metrics (BERTScore: 0.9656; SimCSE: 0.9523), indicating that such similarity metrics failed to detect the crucial change in clinical meaning between "not throat" and "not so".

Example B (Table \ref{tab:qual_example_b}) demonstrates the inverse case of a transcription with elevated WER (0.3333), yet \textbf{No Clinical Impact (0)}. Despite the ASR substituting "sitting" for "feeling", core clinical meaning was preserved given the conversational context. Standard context-agnostic metrics, however, penalized the ASR transcript (Table~\ref{tab:metrics_comparison}), unable to recognize that the core clinical information remained intact. 
In contrast to existing automated metrics, the LLM-as-a-Judge predictions aligned with the expert clinicians' assessment on both examples.

\vspace{1em}
\noindent\textbf{Example A - Significant Clinical Impact, Low WER}

% --- 1. THE DIALOGUE TABLE ---
\begin{table}[H]
\centering
% Defined columns to be slightly wider (0.47) to maximize space usage
\begin{tabular}{p{0.47\linewidth} p{0.47\linewidth}}
\toprule
% --- SHARED HEADER ---
\multicolumn{2}{c}{\textbf{Context}} \\
\midrule

% --- SHARED CONTEXT ROW ---
% We use p{0.96\linewidth} to span the width of both columns plus the gap.
\multicolumn{2}{p{0.96\linewidth}}{
(6) Doctor: Mm-hmm. \par
\vspace{0.3em} % Slight spacing between speakers
(6) Patient: And now I'm, um, just starting to notice a few things. Yeah, I'm having quite shallow breath. \par
\vspace{0.3em}
(7) Doctor: You're having difficulties breathing since then, yeah? OK. So, is there any swelling of your lips, you feel your throat is closing up?
} \\
\midrule

% --- SPLIT HEADERS ---
\textbf{Ground Truth} & \textbf{Transcript} \\
\midrule

% --- DIFFERING CONTENT ---
\textbf{(7) Patient: Not throat, but I can , yeah, I can I can definitely feel something in the lips, yeah.} 
& 
\textbf{(7) Patient: not so but i can i yeah i can i can definitely feel something in the lips yeah} \\
\bottomrule
\end{tabular}
\caption{Example A: High-risk error (Significant Impact) where the crucial word "throat" is substituted for "so".}
\label{tab:qual_example_a}
\end{table}

% --- 2. CLINICIAN REASONING ---
\begin{tcolorbox}[colback=gray!10, colframe=gray!50, title=\textbf{Justification}, breakable]
\small
\textbf{Clinician:} We don't know if "not so" refers to "not throat" - "not so" could be breathing even. \\
\textbf{LLM-as-a-Judge:} The transcription incorrectly captures the patient's response to a critical question about symptoms of a severe allergic reaction. The doctor asks about lip swelling and the throat closing up. In the ground truth, the patient explicitly denies throat symptoms ("Not throat") while confirming lip symptoms. This is a crucial distinction, as throat closing (laryngeal edema) is a sign of a life-threatening emergency (anaphylaxis) requiring immediate intervention. The transcription replaces the clear denial "Not throat" with the ambiguous phrase "not so." A clinician reading "not so" might interpret it as "not really" or "not that much," which fails to definitively rule out the dangerous symptom of throat involvement. This ambiguity in the context of a patient reporting "shallow breath" could lead to a misassessment of the severity of the reaction and potentially delay or alter the appropriate emergency treatment.
\end{tcolorbox}

\vspace{1em}
\noindent\textbf{Example B - No Change In Understanding, Elevated WER}

% --- 1. THE DIALOGUE TABLE ---
\begin{table}[H]
\centering
% Defined columns to be slightly wider (0.47) to maximize space usage
\begin{tabular}{p{0.47\linewidth} p{0.47\linewidth}}
\toprule
% --- SHARED HEADER ---
\multicolumn{2}{c}{\textbf{Context}} \\
\midrule

% --- SHARED CONTEXT ROW ---
% We use p{0.96\linewidth} to span the width of both columns plus the gap.
\multicolumn{2}{p{0.96\linewidth}}{
(9) Doctor: Any abnormal smells or tastes? \par
\vspace{0.3em} % Slight spacing between speakers
(9) Patient: No. \par
\vspace{0.3em}
(10) Doctor: Any, any nausea, vomiting?
} \\
\midrule

% --- SPLIT HEADERS ---
\textbf{Ground Truth} & \textbf{Transcript} \\
\midrule

% --- DIFFERING CONTENT ---
\textbf{(10) Patient: Uh, no, no, been feeling fine actually.} 
& 
\textbf{(10) Patient: no no it's sitting fine actually} \\
\bottomrule
\end{tabular}
\caption{Example B: Benign error (No Clinical Impact) with elevated WER. Although the transcript discusses "sitting" rather than "feeling", the core clinical meaning is contextually preserved in context of the doctor's question.}
\label{tab:qual_example_b}
\end{table}

% --- 2. CLINICIAN REASONING ---
\begin{tcolorbox}[colback=gray!10, colframe=gray!50, title=\textbf{Justification}, breakable]
\small
\textbf{Clinician:} Same understanding\\
\textbf{LLM-as-a-Judge:} The transcription incorrectly captured the patient's response to the question about nausea and vomiting. The ground truth is "been feeling fine actually," while the transcription states, "it's sitting fine actually." In the context of a gastrointestinal review of systems, both phrases convey the same clinical meaning: the patient is denying symptoms of nausea and vomiting. A clinician would interpret "it's sitting fine" as a colloquial way of saying they are not experiencing GI upset. The core clinical information (absence of symptoms) is preserved, making this a cosmetic error with no impact on clinical understanding.
\end{tcolorbox}

\clearpage

% --- 3. THE METRICS TABLE FOR BOTH EXAMPLES ---
% This table shows how the metrics failed to flag this as a major error (e.g., high BLEU, Low WER)
% DATA Manually Checked

\vspace{1em}
\noindent\textbf{Evaluation Metrics Data for Both Examples}

\begin{table}[H]
\centering
\small
\begin{tabular}{lccc}
\toprule
 & Example A & Example B & Risk-Aligned? \\
\midrule
WER $\downarrow$ & 0.1176 & 0.3333 & No \\
CER $\downarrow$ & 0.0875 & 0.2500 & No \\
MER $\downarrow$ & 0.1111 & 0.3333 & No \\
WIL $\downarrow$ & 0.1634 & 0.5556 & No \\
S-WER $\downarrow$ & 0.0094 & 0.1111 & No \\
\midrule
BLEU-1 $\uparrow$ & 0.8889 & 0.6667 & No \\
BLEU-2 $\uparrow$ & 0.8245 & 0.5164 & No \\
BLEU-3 $\uparrow$ & 0.7760 & 0.0000 & No \\
BLEU-4 $\uparrow$ & 0.7277 & 0.0000 & No \\
ROUGE-L $\uparrow$ & 0.9143 & 0.6667 & No \\
ROUGE-1 $\uparrow$ & 0.9143 & 0.6667 & No \\
ROUGE-2 $\uparrow$ & 0.7879 & 0.4000 & No \\
ROUGE-W $\uparrow$ & 0.9143 & 0.6667 & No \\
chrF $\uparrow$ & 0.8387 & 0.5767 & No \\
chrF++ $\uparrow$ & 0.8459 & 0.5659 & No \\
METEOR $\uparrow$ & 0.9214 & 0.6250 & No \\
\midrule
SeMaScore $\uparrow$ & 0.8813 & 0.6849 & No \\
Intelligibility $\uparrow$ & 0.8670 & 0.6151 & No \\
HEVAL $\downarrow$ & 0.0059 & 0.0602 & No \\
Clinical BLEURT $\uparrow$ & 0.4967 & 0.0982 & No \\
BLEURT $\uparrow$ & 0.3749 & 0.1722 & No \\
BARTScore $\uparrow$ & -2.9802 & -3.9300 & No \\
SBERT Sim $\uparrow$ & 0.7402 & 0.5003 & No \\
NLI (XSmall) $\uparrow$ & 0.8863 & 0.1430 & No \\
NLI (Base) $\uparrow$ & 0.9963 & 0.9743 & No \\
NLI (Large) $\uparrow$ & 0.9680 & 0.0433 & No \\
BERTScore $\uparrow$ & 0.9656 & 0.8942 & No \\
SimCSE $\uparrow$ & 0.9523 & 0.5814 & No \\
\midrule
LLM-as-a-Judge Prediction $\downarrow$ & 2 & 0 & Yes \\
\midrule
Reconciled Clinician Label $\downarrow$ & 2 & 0 & N/A \\
\bottomrule
\end{tabular}
\caption{Existing metric scores for Example A (Significant Impact) and Example B (No Impact), with LLM-as-a-Judge predictions and clinician labels. The 'Risk-Aligned?' column indicates whether the metric correctly identifies A as a more clinically impactful error (lower quality transcript) than B. Arrows indicate the direction of improvement for a metric ($\uparrow$ higher is better quality; $\downarrow$ lower is better quality).}
\label{tab:metrics_comparison}
\end{table}

\clearpage

\twocolumn

\section{LLM-as-a-Judge Automation}
\subsection{Model Comparison}
\label{appendix:llm_judge_comparison}
This section provides a detailed comparison of the performance of various LLMs on the clinical impact classification task, using the final GEPA-optimized prompt. All evaluations were conducted on the 50-item held-out test set, with results averaged over five independent runs. The results demonstrate that while the optimized prompt is effective across models, its performance is maximized by Gemini-2.5-Pro, particularly on the most clinically nuanced classification tasks.

\begin{figure}[h!]
  \centering
  \includegraphics[width=\columnwidth]{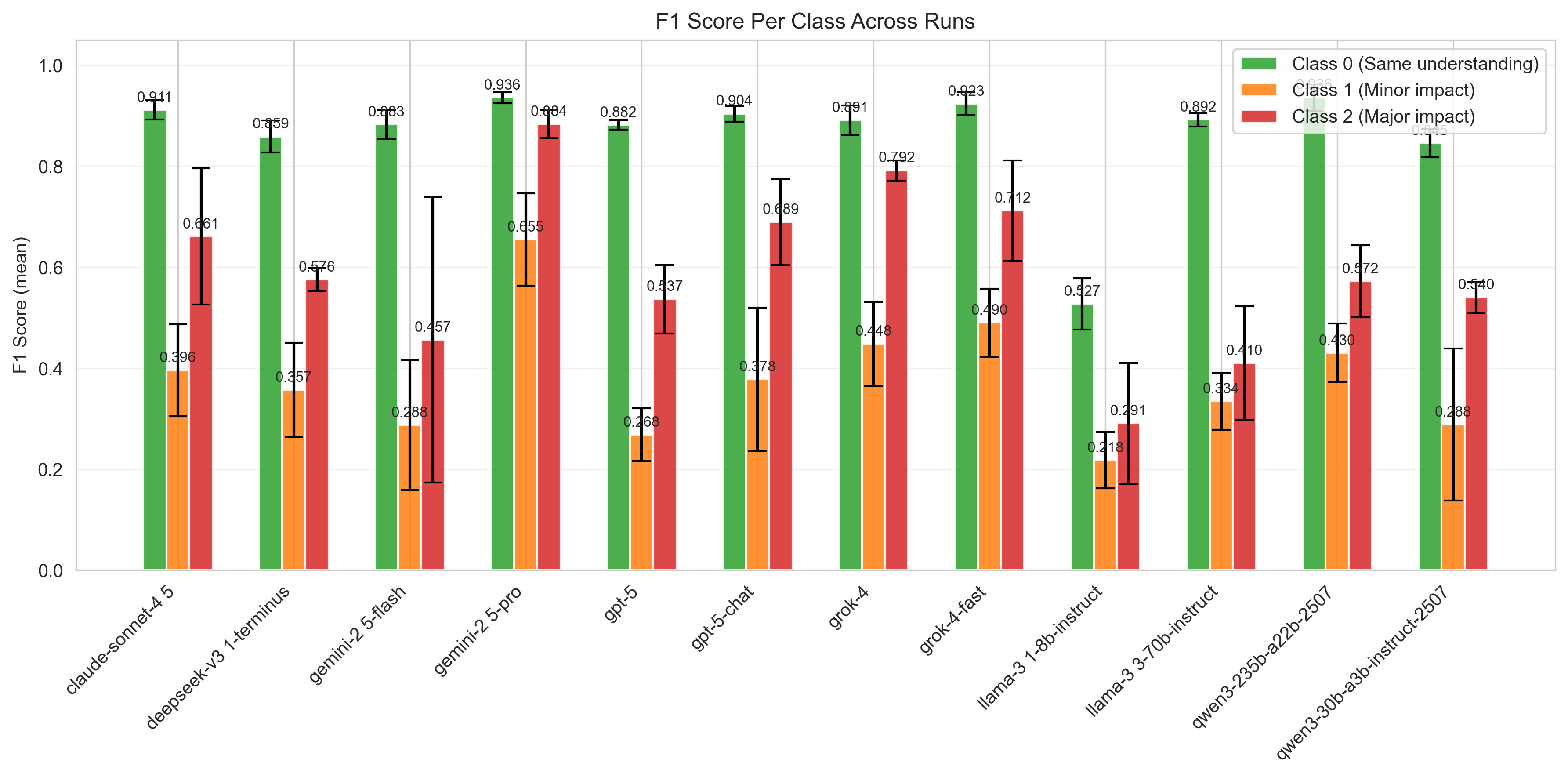}
  \caption{While most high-performing models can reliably identify 'No Impact' (Class 0) and 'Significant Impact' (Class 2) errors, they struggle with the nuanced 'Minimal Impact' (Class 1) category. This highlights the difficulty of discerning subtle changes in clinical meaning, a task where Gemini-2.5-Pro demonstrates unique proficiency as the only model to achieve an F1 score over 0.5, where it got 0.655, for this challenging class. Error bars represent standard deviation across 5 runs.}
  \label{fig:run_f1_score_per_class}
\end{figure}

\begin{table}[h!]
    \centering
    \resizebox{\columnwidth}{!}{%
    \begin{tabular}{lccc}
    \hline
    \textbf{Model} & \textbf{Provider} & \textbf{Macro F1} & \textbf{Cohens $\kappa$} \\
    \hline
    \textbf{gemini-2.5-pro} & Vertex AI & $0.825 \pm 0.000707$ & $0.790 \pm 0.000472$ \\
    grok-4 & xAI & $0.710 \pm 0.001448$ & $0.638 \pm 0.003150$ \\
    grok-4-fast & xAI & $0.708 \pm 0.002754$ & $0.645 \pm 0.003754$ \\
    gpt-5-chat & OpenAI & $0.657 \pm 0.003607$ & $0.588 \pm 0.004064$ \\
    claude-sonnet-4.5 & Anthropic & $0.656 \pm 0.006194$ & $0.589 \pm 0.010008$ \\
    qwen3-235b-a22b-2507 & Nebius AI & $0.646 \pm 0.001014$ & $0.592 \pm 0.001982$ \\
    gpt-5 & OpenAI & $0.562 \pm 0.000805$ & $0.459 \pm 0.000790$ \\
    qwen3-30b-a3b-instruct-2507 & Nebius AI & $0.558 \pm 0.003491$ & $0.428 \pm 0.005847$ \\
    llama-3.3-70b-instruct & Crusoe & $0.545 \pm 0.002365$ & $0.451 \pm 0.002753$ \\
    gemini-2.5-flash & Vertex AI & $0.542 \pm 0.021385$ & $0.450 \pm 0.032153$ \\
    llama-3.1-8b-instruct & Groq & $0.345 \pm 0.001609$ & $0.138 \pm 0.001388$ \\
    \hline
    \end{tabular}
    }
    \caption{The table details the aggregate performance of each LLM judge. The data shows a consistent trend across both metrics (F1-score and Cohen's $\kappa$), with Gemini-2.5-Pro establishing a significant lead. Results are presented as Mean ± Standard Deviation over 5 runs.}
    \label{tab:llm-comparison-full}
\end{table}

\subsection{Clinicians and Judge Agreement}
% Table of agreement between clinicians and judge
\begin{table}[H]
    \centering
    \resizebox{\columnwidth}{!}{%
    \begin{tabular}{lccc}
    \hline
    & \textbf{Clinician A} & \textbf{Clinician B} & \textbf{Judge} \\
    \hline
    \textbf{Clinician A} & --- & 0.505 (0.285, 0.708) & \textbf{0.713 (0.535, 0.867)} \\
    \textbf{Clinician B} & 0.505 (0.285, 0.708) & --- & 0.497 (0.273, 0.702) \\
    \textbf{Judge} & \textbf{0.713 (0.535, 0.867)} & 0.497 (0.273, 0.702) & --- \\\hline
    \end{tabular}
    }
    \caption{Agreement between clinicians and judge using Cohen's $\kappa$ with 95\% bootstrap confidence intervals.}
    \label{tab:iaa_with_judge}
\end{table}

\end{document}